\pdfoutput=1

\documentclass[11pt]{article}

\usepackage[final]{acl}

\usepackage{times}
\usepackage{latexsym}

\usepackage[T1]{fontenc}

\usepackage[utf8]{inputenc}

\usepackage{microtype}

\usepackage{inconsolata}

\usepackage{graphicx}

\usepackage{multirow} 
\usepackage{array}
\usepackage{booktabs}
\usepackage{float}
\usepackage{amsmath}
\usepackage{caption}
\usepackage{subcaption}
\usepackage{algorithm}
\usepackage[noend]{algpseudocode}
\usepackage{enumitem}
\usepackage{amssymb}
\usepackage{nicefrac}
\usepackage{mathtools}
\usepackage{tablefootnote}
\usepackage{ragged2e}
\usepackage{makecell}
\usepackage{calc}

\title{Self-Critique and Refinement for Faithful Natural Language Explanations}

\author{Yingming Wang \quad Pepa Atanasova \\
        \\
        University of Copenhagen\\
        \href{mailto:yiwa@di.ku.dk}{yiwa@di.ku.dk} \quad \href{mailto:pepa@di.ku.dk}{pepa@di.ku.dk}
}

\begin{document}
\maketitle
\begin{abstract}
\label{sec:abstract}
With the rapid development of Large Language Models (LLMs), Natural Language Explanations (NLEs) have become increasingly important for understanding model predictions. However, these explanations often fail to faithfully represent the model's actual reasoning process. While existing work has demonstrated that LLMs can self-critique and refine their initial outputs for various tasks, this capability remains unexplored for improving explanation faithfulness. To address this gap, we introduce Self-critique and Refinement for Natural Language Explanations (SR-NLE), a framework that enables models to improve the faithfulness of their own explanations -- specifically, post-hoc NLEs -- through an iterative critique and refinement process without external supervision. Our framework leverages different feedback mechanisms to guide the refinement process, including natural language self-feedback and, notably, a novel feedback approach based on feature attribution that highlights important input words. Our experiments across three datasets and four state-of-the-art LLMs demonstrate that SR-NLE significantly reduces unfaithfulness rates, with our best method achieving an average unfaithfulness rate of 36.02\%, compared to 54.81\% for baseline -- an absolute reduction of 18.79\%. These findings reveal that the investigated LLMs can indeed refine their explanations to better reflect their actual reasoning process, requiring only appropriate guidance through feedback without additional training or fine-tuning. Our code is available at \url{https://github.com/ymwangv/SR-NLE}.
\end{abstract}

\section{Introduction}
\label{sec:introduction}

\begin{figure}[ht]
    \centering
    \includegraphics[width=\linewidth]{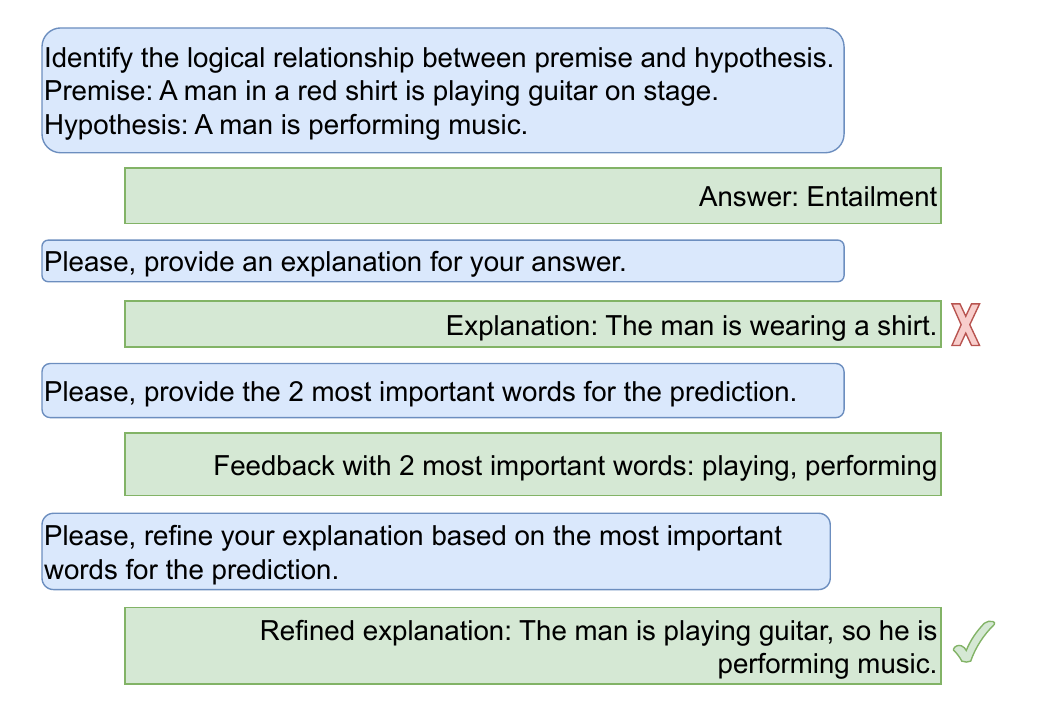}
    \caption{Illustration of our framework SR-NLE improving the faithfulness of the initially generated NLE by providing self-critique of the most important words used in the prediction.}
    \label{fig:example}
\end{figure}

With the rapid development of Large Language Models (LLMs), both closed-source models \citep{openai2024gpt4technicalreport, geminiteam2025geminifamilyhighlycapable} and open-source alternatives \citep{qwen2025qwen25technicalreport, grattafiori2024llama3herdmodels} have demonstrated remarkable capabilities across a wide range of Natural Language Processing (NLP) tasks. Yet, despite these advancements, understanding the reasoning behind their predictions remains a critical challenge -- especially in applications demanding trust and accountability.

Natural Language Explanations (NLEs) have emerged as a promising solution by offering human-readable justifications for model predictions without requiring access to internal model mechanisms. However, ensuring their faithfulness remains a significant challenge. Recent studies have shown that NLEs generated by LLMs often fail to reflect the actual reasoning process of the model \citep{atanasova-etal-2023-faithfulness,  turpin2023language, lanham2023measuringfaithfulnesschainofthoughtreasoning}.

While prior work has primarily relied on changes of the model architecture or additional fine-tuning \cite{yuan2025graphguidedtextualexplanationgeneration,wang2023pinto,atanasova2022diagnostics}, we instead explore whether models possess the capability to independently assess and refine their own explanations. Supporting this direction, recent studies have shown that LLMs are capable of improving their outputs through iterative self-refinement \citep{madaan2023selfrefine, shinn2023reflexionlanguageagentsverbal}. Following this, we ask whether LLMs know if and when their NLEs are faithful to their own internal reasoning by providing self-critique and refining their NLEs for improved faithfulness.

Building on this idea, we propose \textbf{Self-critique and Refinement for Natural Language Explanations (SR-NLE)}, a framework that enables models to improve the faithfulness of their own explanations through an iterative critique and refinement process without external supervision, as only the model itself has access to its internal reasoning and is therefore best positioned to assess explanation faithfulness. Our framework specifically targets \textit{post-hoc NLEs}, where the explanation is generated after the model makes a prediction. Starting from an initial explanation, the model receives feedback identifying potential issues and generates a refined explanation accordingly. This process can be repeated multiple times, enabling incremental improvements.

A central component of SR-NLE is the design of feedback mechanisms that guide the refinement process. We explore two approaches: \emph{natural language feedback (NLF)}, which offers self-critiques in free-form text, and \textbf{a novel feedback mechanism -- \emph{important word feedback (IWF)}}, which identifies important input words for the prediction that are overlooked in the initially generated NLE. For IWF, we implement both prompt-based and attribution-based variants, including attention-based and gradient-based techniques.

We validate the effectiveness of SR-NLE through extensive experiments across three reasoning datasets and four state-of-the-art LLMs, showing consistent improvements in explanation faithfulness over prior methods and strong baselines.

\paragraph{Our main contributions are as follows:}
\begin{itemize}[leftmargin=12pt,itemsep=0pt]
    \item We introduce \textbf{SR-NLE}, a novel framework that enables models to improve the faithfulness of their explanations through iterative self-critique and refinement guided by different feedback mechanisms, without external assistance, architectural modifications or specialized training.
    \item We propose and evaluate \textbf{multiple feedback strategies for faithfulness}, including natural language feedback (NLF) and \textbf{a novel feedback mechanism -- important word feedback (IWF)}, that leverages feature attribution to identify important input words in generated NLEs.
    \item We empirically demonstrate that SR-NLE significantly reduces unfaithfulness rates across multiple datasets and models. Our best method (attention-based IWF) achieves an average unfaithfulness rate of 36.02\% compared to 54.81\% for initial NLEs -- an absolute reduction of 18.79\% unfaithfulness.
\end{itemize}

\section{Related Works}
\label{sec:related}

\paragraph{Natural Language Explanations and Faithfulness Evaluation}  
Natural Language Explanations (NLEs) provide human-readable justifications for model predictions, traditionally obtained via supervised training on annotated datasets \citep{NEURIPS2018_4c7a167b, rajani-etal-2019-explain, atanasova-etal-2020-generating-fact}. Recently, LLMs have enabled NLE generation via in-context learning \citep{brown2020languagemodelsfewshotlearners}. A prominent example is chain-of-thought reasoning \citep{wei2022chain, kojima2022large}, where LLMs generate intermediate reasoning steps alongside the model prediction. \emph{In contrast, our work focuses on post-hoc NLEs, where explanations are generated after the model prediction is made}. Despite these advances in NLE generation, numerous studies have identified a gap between generated NLEs and the model's actual reasoning process \citep{atanasova-etal-2023-faithfulness, turpin2023language, lanham2023measuringfaithfulnesschainofthoughtreasoning}. To quantify this faithfulness gap, researchers have proposed various automatic evaluation metrics, such as counterfactual tests \citep{atanasova-etal-2023-faithfulness, siegel-etal-2024-probabilities} and association-based measures \citep{wiegreffe2022measuringassociationlabelsfreetext, parcalabescu2024measuring}. \emph{In this work, we adopt counterfactual tests \citep{atanasova-etal-2023-faithfulness} as our evaluation method, as they offer instance-level, automatic assessments of explanation faithfulness without requiring human annotations}.

\paragraph{Frameworks for Improving NLE Faithfulness} Existing frameworks for improving the faithfulness of NLEs employ strategies that either make changes to the model architecture or require an additional NLE fine-tuning stage with newly introduced objectives. \citet{pmlr-v162-majumder22a} proposed a knowledge-grounded approach that leverages external commonsense knowledge during fine-tuning to enrich explanations.  \citet{wang2023pinto} introduced a two-stage approach using counterfactual regularization to align predictions with generated explanations. Architectural modifications have shown promise in the state-of-the-art G-Tex framework \cite{yuan2025graphguidedtextualexplanationgeneration}, which encodes highlight explanations via a graph neural network to guide NLE generation. \textit{Our SR-NLE framework distinguishes itself from existing work by enabling models to improve explanation faithfulness through iterative self-critique and refinement -- without external supervision, architectural modifications, or task-specific training -- entirely based on the model’s own internal knowledge.}

\paragraph{Self-Refinement Methods}
Recent work has shown that LLMs can improve their own outputs via iterative self-refinement. The Self-Refine approach \citep{madaan2023selfrefine} demonstrates that models can critique and revise their own outputs, leading to improved performance across a variety of tasks. Similarly, \citet{shinn2023reflexionlanguageagentsverbal} explore self-reflection mechanisms for agent-level reasoning. \textit{While these works establish the general potential of self-improvement, they do not specifically address the challenge of improving explanation faithfulness}. Building on the general idea of self-refinement, Cross-Refine \citep{wang2024crossrefineimprovingnaturallanguage} applies this paradigm to NLE generation. Their framework adopts a \textit{cross-model} design, where one LLM generates the initial explanation and another, separate LLM provides feedback and suggestions for revision. \textit{In contrast, SR-NLE operates entirely within a single model, leveraging its internal capabilities for both critique and refinement. Furthermore, SR-NLE focuses specifically on improving faithfulness using automated counterfactual tests for objective evaluation}. In comparison, Cross-Refine primarily evaluates explanation quality through multiple automated metrics, while relying on human judgments for faithfulness assessment.\looseness=-1

\paragraph{Input Feature Attribution Methods}
Input feature attribution methods quantify how much each input feature contributes to a model's prediction. Common approaches include Shapley values \citep{lundberg2017unifiedapproachinterpretingmodel}, integrated gradients \citep{sundararajan2017axiomaticattributiondeepnetworks}, and attention weights \citep{jain2019attentionexplanation}. A newer paradigm leverages prompt-based approaches \citep{kroeger2023are}, where LLMs are prompted to directly identify influential input features. The most prevalent application of these methods is to provide post-hoc explanations, helping humans understand which parts of the input most strongly influence the model’s decision-making. Beyond interpretability, these methods have also been used to construct rationales for in-context exemplars in few-shot learning to improve task accuracy. AMPLIFY\citep{krishna2023posthocexplanationslanguage} trains a proxy model and applies attribution methods to extract important words, which are then converted into rationales for few-shot exemplars. Self-AMPLIFY\citep{bhan2024selfamplifyimprovingsmalllanguage} extends this idea by removing the proxy and computing attributions directly from LMs to obtain important words, which are likewise used as rationales for exemplar construction. In the context of NLEs, G-Tex \citep{yuan2025graphguidedtextualexplanationgeneration} leverages attribution-derived highlights to guide the generation of NLEs through graph encoding. \emph{In this work, we propose a novel use of attribution methods to obtain feedback for improving the faithfulness of LLM-generated NLEs. Similar to SELF-AMPLIFY \citep{bhan2024selfamplifyimprovingsmalllanguage}, we apply attribution methods directly to LLMs to extract important words and use them to construct feedback to guide the model in generating more faithful NLEs.}

\section{Method}
\label{sec:method}

In this section, we present \textbf{SR-NLE}, a framework for improving the faithfulness of NLEs generated by LLMs. SR-NLE employs an iterative self-critique and refinement process, enabling LLMs to progressively identify faithfulness issues in their own NLEs and make targeted improvements thereof. This framework leverages the in-context learning and self-improvement capabilities of LLMs, without requiring human involvement or additional models for feedback.

\subsection{Preliminary}
\label{sec:preliminary}
The SR-NLE framework operates on the assumption that LLMs have the capability to identify and improve their own explanations when guided with appropriate prompts. Our framework relies entirely on a single model $\mathcal{M}$ for all components, without requiring human involvement or additional models. For an input~$x$, the model first predicts an answer $y$ and produces an initial explanation $e^{0}$, then, through an iterative process of self-critique and refinement, after each round $r$, we obtain a progressively improved explanation $e^{r}$. To direct the model in different stages of the framework, we employ four categories of prompts: $p_{\text{ans}}$ for answer generation, $p_{\text{exp}}$ for explanation generation, $p_{\text{fb}}$ for feedback generation, and $p_{\text{ref}}$ for refinement generation, where both feedback and refinement prompts have two variants corresponding to our two feedback approaches: natural language feedback (NLF) and important word feedback (IWF). Throughout the framework, we use ``$\oplus$'' to denote filling a prompt template with its variables.

\begin{figure}[t]
    \centering
    \includegraphics[width=1.0\linewidth]{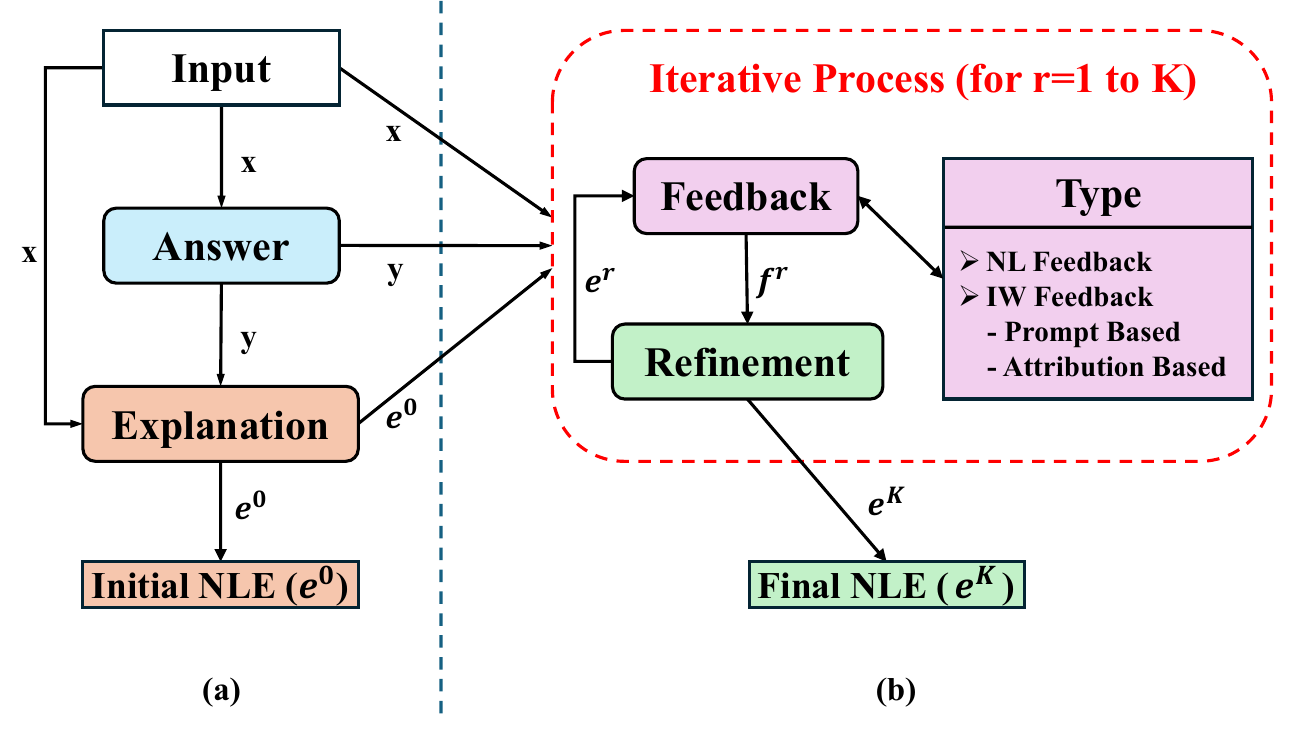}
    \caption{The SR-NLE framework. (a) Answer and Explanation Generation Phase: The framework produces the answer $x$ and initial explanation $e^{0}$. (b) Iterative Critique and Refinement Phase: The framework iteratively improves explanations through feedback-refinement loops over multiple rounds.}
    \label{fig:framework}
\end{figure}
\subsection{SR-NLE Framework}
\label{sec:framework}
Our SR-NLE framework consists of two main phases: (a) Answer and Explanation Generation, which produces the answer and its initial explanation, and (b) Iterative Critique and Refinement, which progressively improves the explanation through multiple rounds. Figure~\ref{fig:framework} illustrates this two-phase process. The algorithmic formulation is provided as Algorithm~\ref{alg:srnle}.

\subsubsection{Answer and Explanation Generation}
This phase produces the answer and its initial explanation (see Figure~\ref{fig:framework}a):

\paragraph{Answer.} 
Given an input $x$, the model first generates an answer:
\begin{equation}
  y = \mathcal{M} \bigl(p_{\text{ans}} \oplus x \bigr)
  \label{eq:ans}
\end{equation}

\paragraph{Explanation.} 
Using the answer $y$, the model generates an initial explanation:
\begin{equation}
  e^{0} = \mathcal{M} \bigl(p_{\text{exp}} \oplus x \oplus y\bigr)
  \label{eq:exp}
\end{equation}
This initial explanation serves as the starting point for our iterative refinement process.

\subsubsection{Iterative Critique and Refinement}
This phase forms the core of our framework, where explanations are iteratively improved for $K$ rounds (see Figure~\ref{fig:framework}b):

\paragraph{Feedback.} 
For each refinement round $r$, the model generates feedback on the preceding explanation $e^{r-1}$. We explore two distinct feedback approaches:

\begin{itemize}[leftmargin=12pt,itemsep=-0.5em]
    \item \textbf{Natural Language Feedback (NLF).} With this approach, $\mathcal{M}$ generates detailed textual self-critique for each round $r$:
    \begin{equation}
      f_{\text{nl}}^{r} = \mathcal{M} \bigl(p_{\text{fb}} \oplus x \oplus y \oplus e^{r-1} \bigr)
      \label{eq:fb_nl}
    \end{equation}
    
    \item \textbf{Important Word Feedback (IWF).} This novel feedback approach leverages attribution explanations, which mark specific input tokens~\cite{deyoung-etal-2020-eraser} or segments~\cite{choudhury-etal-2023-explaining} critical to a model's prediction. While these explanations may lack the plausibility of NLEs~\citep{jie2024interpretable}, their faithfulness is straightforward to measure and has seen significant improvements~\citep{sun-etal-2025-evaluating,atanasova-etal-2020-diagnostic}. We hypothesise that such explanations can enhance NLE faithfulness by providing explicit feedback about which input elements should be emphasized in the generated explanation. Our approach identifies words in the input that are most important for the answer:
    \begin{equation}
        \begin{aligned}
        &&  \mathcal{S} = \textsc{Score} \bigl(x, y \bigr) && \\
        && \mathcal{I} = \textsc{Select} \bigl( \mathcal{S}, N \bigr) && \\
        && f_{\text{iw}} = \textsc{Format} \bigl( \mathcal{I} \bigr) &&
        \end{aligned}
    \label{eq:fb_iw}
    \end{equation}
    
    Here, we employ a method $\textsc{Score}$ to provide a list $\mathcal{S}$ of the words in input $x$ with their importance scores for answer $y$. From these scored words, we select the top-$N$ most important ones -- $\mathcal{I}$, to form the feedback. We implement two $\textsc{Score}$ methods:
    \begin{itemize}[leftmargin=12pt,itemsep=-0.5em]
        \item \textbf{Prompt-based}: Following \citet{kroeger2023are}, who find that LLMs can be used with high accuracy as post-hoc explainers, we prompt the model itself to assign importance scores to input words (IWF-Pmt):
        \begin{equation}
        \textsc{Score} = \mathcal{M}(p_{\text{fb}} \oplus x \oplus y)
        \end{equation}
        \item \textbf{Attribution-based}: Following \citet{bhan2024selfamplifyimprovingsmalllanguage}, we use input feature attribution methods to quantify and assign importance scores to words (IWF-Attr). We detail our method for computing the IWF-Attr \textsc{Score} in Section~\ref{sec:attr_iwf}.
    \end{itemize}
\end{itemize}

\paragraph{Refinement.}
Using the feedback, the model refines its explanation:
\begin{equation}
    e^{r} = \mathcal{M} \bigl( p_{\text{ref}} \oplus x \oplus y \oplus e^{r-1} \oplus f^{\star} \bigr)
    \label{eq:refine}
\end{equation}
where $f^{\star}$ is either $f_{\text{nl}}^{r}$ or $f_{\text{iw}}$ depending on the feedback type. 

This process of feedback generation and refinement repeats for $K$ rounds, with each round potentially addressing different sources of unfaithfulness in the explanation. After the final round, we obtain $e^K$ as our final NLE.

\subsection{Attribution-Based IWF \textsc{Score}}
\label{sec:attr_iwf}

\begin{figure}[t]
    \centering
    \includegraphics[width=\linewidth]{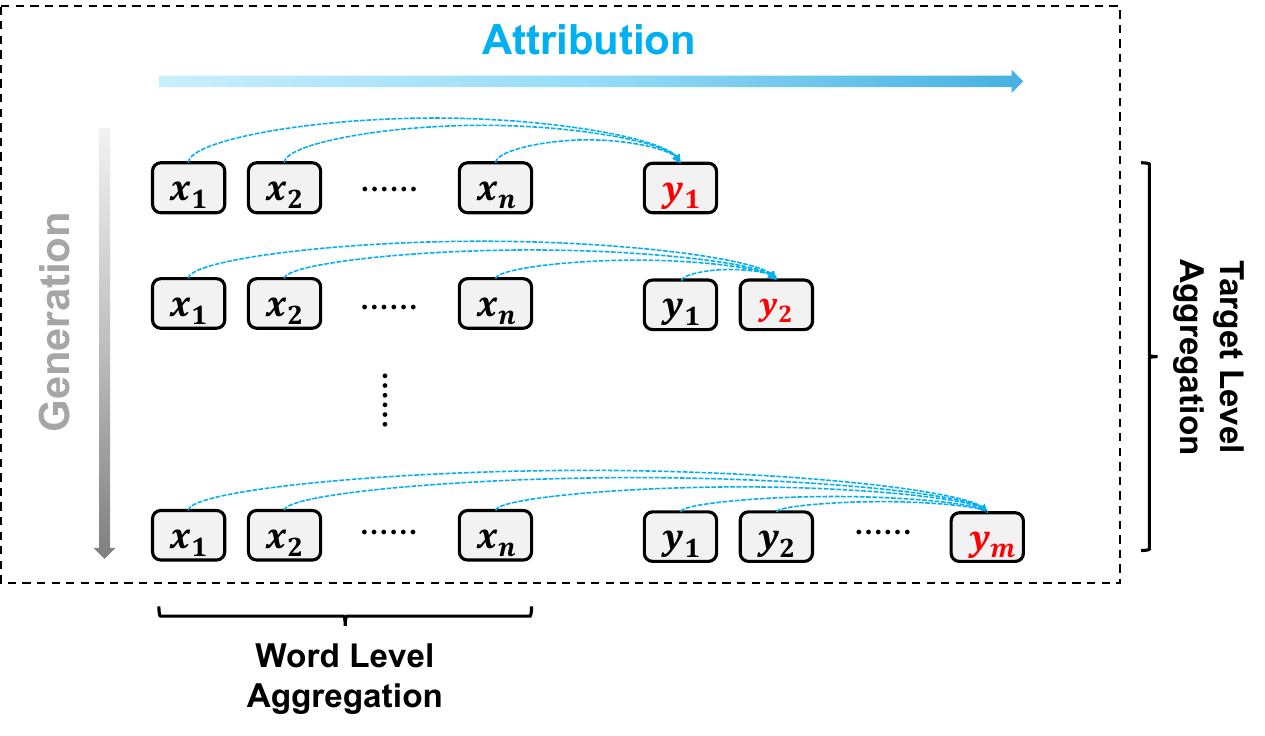}
    \caption{Illustration of attribution-based IWF \textsc{Score}.}
    \label{fig:attr_iwf}
\end{figure}

While prompt-based IWF directly prompts the model to assign importance scores to input words, attribution-based IWF computes these scores using feature attribution methods. Our approach for computing the attribution-based IWF \textsc{Score} is illustrated in Figure~\ref{fig:attr_iwf} and detailed in Algorithm~\ref{alg:attr_iwf}, consisting of the following steps:

\paragraph{Target Span Identification} 
First, we identify the answer span within the model output. Given a task input $x$ and answer generation prompt template $p_{ans}$, we construct the full model input $p_{ans} \oplus x$. After running the model on this combined input, we locate the answer span $y$ within the model output.

\paragraph{Sequential Token Attribution} 
For each token $y_j$ in the answer span, we compute attribution scores considering the entire context available at generation time. This includes all tokens in the full model input, as well as all previously generated tokens.\looseness=-1

\paragraph{Token-level Computation}
We quantify how each token in the full model input contributes to generating each token in the model output:\looseness=-1
\begin{equation}
    a_{i,j} = |\text{Attribution}(x_i, y_j | \text{context}_{<j})|
\end{equation}
where $a_{i,j}$ represents the attribution score of token $x_i$ from the full model input (prompt + task input) for the prediction of output token $y_j$ given all preceding context. We apply the absolute value function for two key reasons: (1) to focus on the magnitude of influence rather than its direction, as both strong positive and negative influences indicate important tokens; and (2) to prevent positive and negative attributions from cancelling each other out during aggregation steps.

\paragraph{Target-level Aggregation} 
We aggregate the token-level attributions across the answer span for each input token:
\begin{equation}
    a_i = \sum\nolimits_{j=1}^{|y|} a_{i,j}
\end{equation}
where we sum (rather than average) the attribution scores to capture the total influence of each input token.\looseness=-1

\paragraph{Word-level Aggregation} 
To obtain word-level importance, we map token attributions back to the original words in the task input (excluding prompt tokens). For words split into multiple tokens during tokenization, we combine their attribution scores:
\begin{equation}
    \text{score}(w) = \sum\nolimits_{i \in \text{indices}(w)} a_i
\end{equation}
where $\text{indices}(w)$ represents the indices of all tokens corresponding to word $w$ in the task input.

\section{Experiments}
\label{sec:experiments}

\subsection{Datasets}
\label{sec:datasets}

We conducted our experiments on three widely used natural language reasoning datasets with NLEs: \textbf{ComVE} \cite{wang-etal-2020-semeval}, \textbf{ECQA} \cite{aggarwal-etal-2021-explanations}, and \textbf{e-SNLI} \cite{NEURIPS2018_4c7a167b}. The task of ComVE is to identify which of the two sentences violates common sense. The task of ECQA is to answer multiple-choice questions requiring common sense reasoning. The task of e-SNLI is to determine the logical relationship (contradiction, neutral or entailment) between the premise and hypothesis. We selected 1,000 instances from each dataset for our experiments due to computational constraints. Details about dataset selection and characteristics are provided in Appendix~\ref{appendix:datasets}.\looseness=-1

\subsection{Models}
\label{sec:models}

We utilized four state-of-the-art open-source models for our experiments: \textbf{Llama} \cite{grattafiori2024llama3herdmodels}, \textbf{Mistral} \cite{jiang2023mistral7b}, \textbf{Qwen} \cite{qwen2025qwen25technicalreport}, and \textbf{Falcon} \cite{almazrouei2023falconseriesopenlanguage}. For each model, we selected its instruction-tuned version, as our framework primarily operates in a zero-shot setting, which relies heavily on the model's ability to follow instructions effectively. Additionally, we limited our selection to models with sizes under 10B parameters to balance performance and computational efficiency. Detailed model specifications are provided in Appendix~\ref{appendix:models}.

\subsection{Evaluation}
\label{sec:evaluation}

To evaluate the faithfulness of the model-generated NLEs, we employ the counterfactual test proposed by \citet{atanasova-etal-2023-faithfulness}. The counterfactual test works by making an intervention to the original instance to get an intervened instance. The evaluation then consists of two steps: (1) Identify counter instances: intervened instances whose prediction changes compared to the original instance. (2) Identify unfaithful instances: counter instances whose NLEs (generated by baseline methods or SR-NLE) do not contain the intervened word (determined by string matching). The unfaithfulness rate is calculated as:
\begin{equation}
\text{Unfaithfulness} = \nicefrac{N_{\text{unfaithful}}}{N_{\text{counter}}}
\end{equation}
This metric allows us to directly compare the faithfulness of NLEs generated by different methods, with lower rates of unfaithfulness being more desirable. We apply this metric consistently across all baseline methods and SR-NLE variants to ensure fair comparison.

\paragraph{Intervention Generation.} 
In our implementation, we adopt the random approach from \citet{atanasova-etal-2023-faithfulness}. Specifically, we randomly select a noun or a verb from any position in the input. For nouns, we prepend a random adjective, and for verbs, we prepend a random adverb. Different from \citet{atanasova-etal-2023-faithfulness}, we further employ prompting GPT-4o \cite{openai2024gpt4technicalreport}, to ensure the generation of multiple effective, coherent, and meaningful interventions for the same instance without duplications. We generate 20 unique interventions for each original instance from each dataset. The detailed intervention generation prompt and quality checks are described in Appendix~\ref{appendix:evaluation}.

\subsection{Baselines}
\label{sec:baselines}

We compare our SR-NLE framework against two baselines suggested by us, as well as an existing prior method:

\paragraph{Init-NLE.} The initial NLEs were generated by the model without any refinement process. This corresponds to $e^0$ in our framework and represents the typical approach used in most NLE generation scenarios.

\paragraph{SC-NLE.} NLEs generated using the Self-Consistency method \cite{wang2023selfconsistency}, where we sample multiple explanations with temperature sampling and select the most representative explanation using the semantic centroid voting (Algorithm~\ref{alg:semantic_centroid_voting}). This approach encodes all candidates using SentenceBERT \cite{reimers2019sentencebertsentenceembeddingsusing}, computes their centroid in the embedding space, and selects the explanation with the highest cosine similarity to this centroid. This effectively identifies the explanation that best represents the consensus meaning across all samples. This baseline represents a strong ensemble-based alternative that does not require iterative refinement. The specific configuration of sampling parameters is discussed in Section~\ref{sec:experimental_setups}.

\paragraph{Comparisons to Prior Work.}
We also compare our SR-NLE with G-TEX~\citep{yuan2025graphguidedtextualexplanationgeneration}, a recent state-of-the-art method that also aims to improve explanation faithfulness. While we do not implement their approach, we report their results from the original paper for reference.

\subsection{Experimental Setups}
\label{sec:experimental_setups}

\paragraph{Implementation Details}
We use greedy decoding throughout our pipeline and experiment with up to $K=3$ refinement rounds. For attribution-based IWF, we compare two attribution methods: (1) gradient-based attribution using Integrated Gradients (IWF-IG; \citet{sundararajan2017axiomaticattributiondeepnetworks}), identified as the most faithful post-hoc explanations \cite{atanasova-etal-2020-diagnostic}, and (2) attention-based attribution (IWF-Attn) leveraging the model's attention mechanisms. A more detailed description of these attribution methods is provided in Appendix~\ref{appendix:attribution}. For all important word feedback variants, we use the top-5 important words as feedback. For the SC-NLE baseline, we sample 20 candidate explanations with temperature 1.0 and select the most representative one using semantic centroid voting, as described in Section~\ref{sec:baselines}. Detailed ablation studies on various parameters are provided in Appendix~\ref{appendix:ablations}.

\paragraph{Prompts.} Our entire pipeline operates in a zero-shot setting, with stage-specific instructions designed for each dataset. Complete prompt templates are provided in Appendix~\ref{appendix:prompt}.

\begin{table*}[ht]
\centering
\renewcommand{\arraystretch}{1.4}
\setlength{\tabcolsep}{2.1pt}
\small                      
\begin{tabular*}{\linewidth}{@{\extracolsep{\fill}}l@{\hspace{4pt}}l|cccc|cccc|cccc|c@{}}
\toprule
\multicolumn{2}{@{}c|}{\multirow{2}{*}{\textbf{Method}}} & \multicolumn{4}{c|}{\textbf{ComVE}} & \multicolumn{4}{c|}{\textbf{ECQA}} & \multicolumn{4}{c|}{\textbf{e-SNLI}} & \multirow{2}{*}{\textbf{Avg.}} \\
\cmidrule{3-14}
\multicolumn{2}{@{}c|}{} & \textbf{Falcon} & \textbf{Llama} & \textbf{Mistral} & \textbf{Qwen} & \textbf{Falcon} & \textbf{Llama} & \textbf{Mistral} & \textbf{Qwen} & \textbf{Falcon} & \textbf{Llama} & \textbf{Mistral} & \textbf{Qwen} & \\

\midrule
\multirow{2}{*}{\textbf{Baseline}} 
& Init-NLE & 69.64 & 72.91 & 70.33 & 69.74 & 49.54 & 42.02 & 47.03 & 52.74 & 22.44 & 59.53 & 58.93 & 42.90 & 54.81\\
& SC-NLE & 63.27 & 71.78 & 63.93 & 68.42 & 44.69 & 39.25 & 44.17 & 51.21 & 19.24 & 43.99 & 47.98 & 38.91 & 49.74\\

\midrule
\multirow{4}{*}{\textbf{SR-NLE}} 
& NLF & 60.71 & 63.67 & 64.29 & 58.99 & 43.77 & 37.76 & 44.72 & 46.79 & 23.01 & 47.04 & 44.18 & 36.04 & 47.58\\

& IWF-Pmt & \underline{44.13} & 62.70 & \underline{46.08} & \underline{51.97} & \underline{24.82} & \bf 24.32 & 43.12 & 29.26 & 22.01 & \underline{36.16} & \underline{37.80} & 24.43 & 37.23\\

& IWF-Attn & 46.43 & \underline{60.37} & \bf 44.39 & \bf 50.66 & 27.03 & \bf 24.32 & \bf 42.03 & \bf \bf 26.28 & \bf 18.21 & \bf 34.94 & 38.49 & \bf 19.10 & \bf 36.02\\

& IWF-IG & \bf 42.09 & \bf 58.20 & 49.10 & 52.85 & \bf 24.60 & \underline{24.81} & \underline{42.66} & \underline{27.77} & \underline{18.35} & 37.38 & \bf 35.86 & \underline{21.98} & \underline{36.30}\\

\bottomrule

\end{tabular*}
\caption{Main results of SR-NLE framework reporting unfaithfulness rates (\%) after three refinement rounds (R3). Best (lowest) results per dataset-model combination are \textbf{bolded}, second best are \underline{underlined}.}

\label{tab:main_results}
\end{table*}


\section{Results}
\label{sec:results}

\subsection{Main Results}
\label{sec:res}

Table~\ref{tab:main_results} presents our comprehensive evaluation results after 3 refinement rounds. Additional results from intermediate refinement rounds, complementary metrics, additional analysis, and detailed visualizations are provided in Appendix~\ref{appendix:results}.

\paragraph{SR-NLE outperforms baselines} The SR-NLE framework shows superior performance over baseline methods in most experimental settings. Our best implementation (IWF-Attn) reduces unfaithfulness rates by an average of \textbf{18.79\%} compared to Init-NLE and by an average of \textbf{13.72\%} compared to SC-NLE. Even our least effective method (NLF), despite underperforming in isolated cases (e.g., e-SNLI with Falcon), still achieves an average reduction of \textbf{7.23\%} compared to Init-NLE and \textbf{2.16\%} compared to the SC-NLE baseline, demonstrating the \textit{overall effectiveness of our framework}.

\paragraph{IWF outperforms NLF} All three IWF implementations consistently outperform NLF across all experimental settings. On average, IWF-Attn, IWF-IG, and IWF-Pmt achieve \textbf{11.56\%}, \textbf{11.28\%}, and \textbf{10.35\%} lower unfaithfulness rates than NLF, respectively. \textit{This performance gap demonstrates that explicit important word feedback provides more effective guidance for refinement than natural language feedback.}

\paragraph{Comparable performance across IWF variants} 
A notable finding is that prompt-based IWF performs similarly to attribution-based implementations, with average unfaithfulness rates of 36.02\% (IWF-Attn), 36.30\% (IWF-IG), and 37.23\% (IWF-Pmt), differing by only \textbf{1.21\%}. This suggests that the IWF framework is robust to different word selection strategies, with prompt-based methods offering practical advantages in terms of efficiency, accessibility, and reliability (3.75\% hallucination rate; see Appendix~\ref{appendix:hallucination_rate}). To better understand this robustness, we also conducted additional experiments exploring the impact of word selection quality on IWF performance (detailed in Appendix~\ref{appendix:selection_quality}).

\subsection{Comparison with Prior Work.}
\label{sec:comparision}

\begin{table}[t]
\centering
\begin{tabular}{lccc}
\toprule
\textbf{Method} & \textbf{ComVE} & \textbf{ECQA} & \textbf{e-SNLI} \\
\midrule
G-TEX & 87.17 & 43.42 & 33.25 \\
SR-NLE & \textbf{44.39} & \textbf{24.32} & \textbf{18.21} \\
\bottomrule
\end{tabular}
\caption{Comparison of dataset-wise best (lowest) unfaithfulness rates (\%) for G-TEX and SR-NLE. Each entry reports the best performance achieved by each method on the respective dataset.}
\label{tab:comparison_prior}
\end{table}

While the results in Table~\ref{tab:comparison_prior} suggest that SR-NLE substantially outperforms the state-of-the-art G-TEX in terms of explanation faithfulness, this comparison is not fully controlled. The reported numbers correspond to each method's best-performing configuration on each dataset, and differences in counterfactual generation strategies or data splits may influence the outcomes. Nevertheless, the results provide a useful reference point, demonstrating the potential of SR-NLE as a lightweight and effective alternative for improving explanation faithfulness, compared to G-TEX, which requires architectural changes and additional fine-tuning.

\subsection{Detailed Analysis}
\label{sec:analysis}
In this section, we conduct further in-depth analyses to better understand the effectiveness of our SR-NLE framework.

\paragraph{Faithfulness State Transitions}

\begin{figure}[t]
    \centering
    \includegraphics[width=\linewidth]{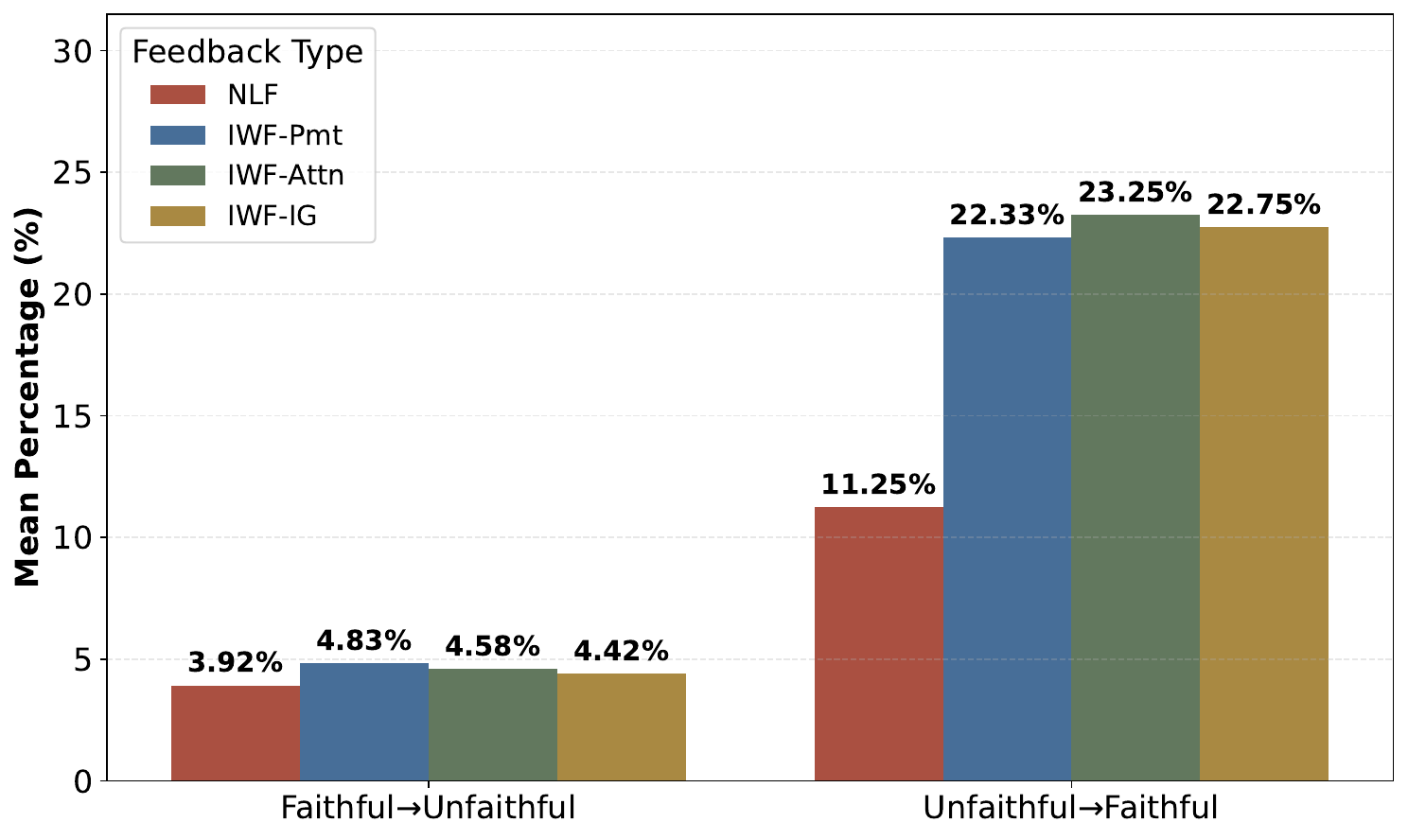}
    \caption{Faithfulness state transitions from $e^0$ to $e^3$ for different feedback methods, averaged across 12 model-dataset combinations. The left group shows the proportion of initially faithful explanations that become unfaithful, while the right group shows the proportion of initially unfaithful explanations that become faithful.}
    \label{fig:state_transitions}
\end{figure}

To understand the refinement mechanism at a granular level, we analyze how individual explanations transition between faithful and unfaithful states. Figure~\ref{fig:state_transitions} presents the transition rates between two key states: faithful→unfaithful (F→U) and unfaithful→faithful (U→F). For all feedback methods, positive transitions (U→F) substantially exceed negative transitions (F→U), with IWF methods showing a particularly favorable ratio. This indicates that our refinement process effectively corrects unfaithful explanations while rarely compromising initially faithful ones. Among all methods, \textit{IWF-Attn achieves the best balance of high positive and low negative transition rates, which explains its lowest overall unfaithfulness rates in our main results}.\looseness=-1

\paragraph{Refinement Efficiency Across Rounds} 

\begin{figure}[t]
    \centering
    \includegraphics[width=\linewidth]{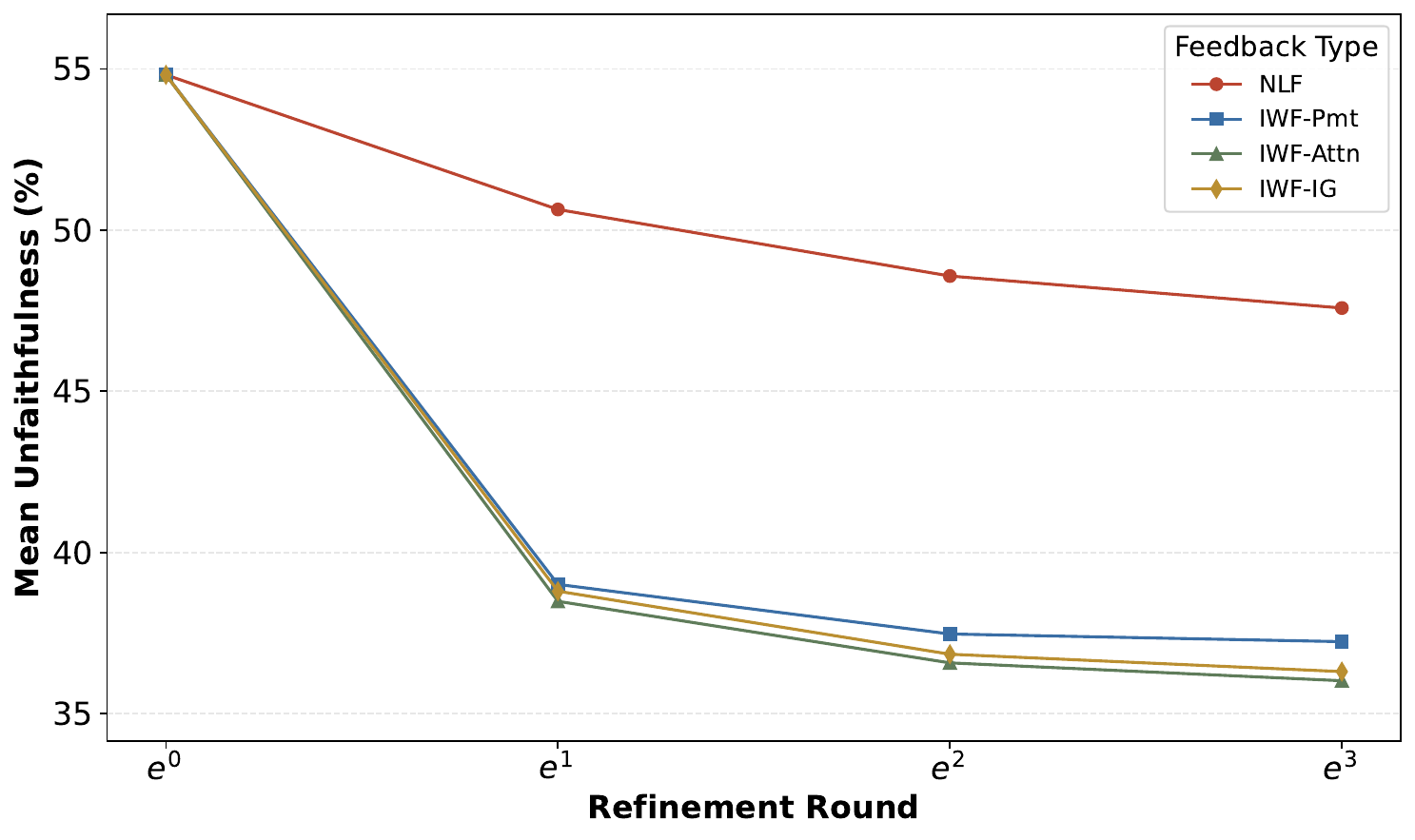}
    \caption{Unfaithfulness rates across successive refinement rounds for feedback methods, averaged across 12 model-dataset combinations.}
    \label{fig:iterative_improvement}
\end{figure}

Figure~\ref{fig:iterative_improvement} illustrates unfaithfulness rates across successive refinement rounds ($e^0$ to $e^3$) for all feedback methods. We can observe two consistent trends: First, unfaithfulness rates continuously decrease with additional refinement rounds, demonstrating the effectiveness of our method. Second, the most substantial reduction occurs during the first refinement round ($e^0$ to $e^1$), with the rate of reduction slowing down in subsequent rounds. These indicate that two or three refinement rounds may offer an optimal trade-off between performance and computational efficiency in practical applications. Therefore, we limited our experiments to a maximum of three refinement rounds.\looseness=-1

\paragraph{Relationship Between Explanation Length and Unfaithfulness Rate}

\begin{figure}[t]
    \centering
    \includegraphics[width=\linewidth]{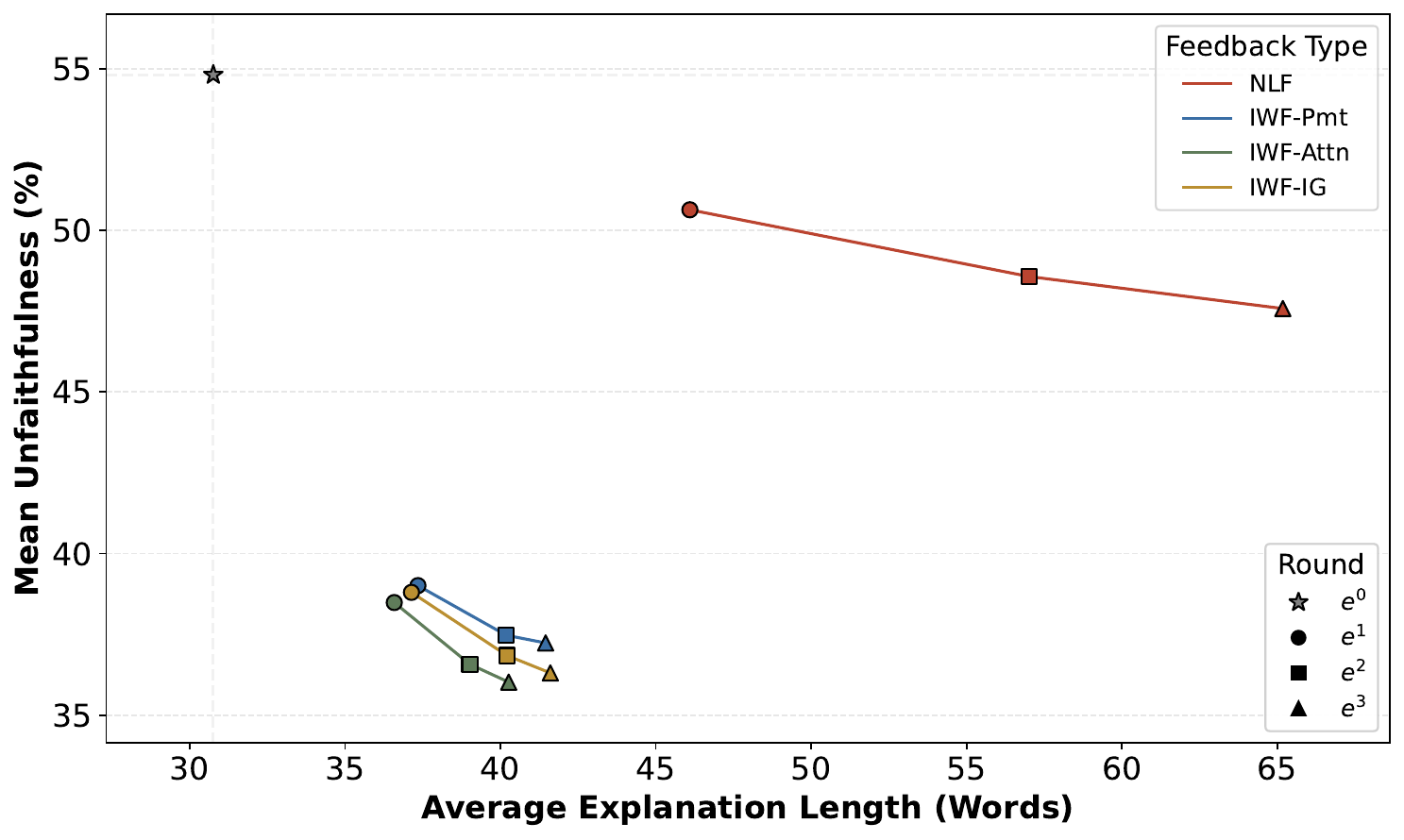}
    \caption{Unfaithfulness rates versus explanation lengths across refinement rounds for feedback methods, averaged across 12 model-dataset combinations.}
    \label{fig:length_unfaithfulness}
\end{figure}

Figure~\ref{fig:length_unfaithfulness} reveals a clear pattern: as explanation length increases through successive refinement rounds, unfaithfulness rates consistently decrease for all feedback methods, indicating a generally inverse relationship between length and unfaithfulness. Initial explanations are the shortest and exhibit the highest unfaithfulness rates, while refined explanations become longer with decreasing unfaithfulness in each refinement round. However, the efficiency of this length-unfaithfulness relationship varies significantly between feedback types. NLF produces substantially longer explanations but achieves relatively modest reductions in unfaithfulness. In contrast, IWF methods reach significantly lower unfaithfulness rates with more moderate length increases. This indicates that although increased length generally reduces unfaithfulness, the focus on important words in the refinement process leads to more efficient reductions. The superior performance of IWF methods demonstrates the effectiveness of our approach in guiding models to address the most relevant aspects of explanations rather than broadly expanding content.\looseness=-1

\begin{table*}[t]
\centering
\small
\begin{tabular}{p{7.5cm}|p{7.5cm}}
\toprule
\multicolumn{2}{l}{
  \begin{minipage}{\dimexpr 15cm + 2\tabcolsep + \arrayrulewidth\relax}
    \begin{tabular}{@{} p{2cm} p{13cm} @{}} 
      \textbf{Original} 
      & \textbf{Question:} There was only one room in the place where Bill slept. It had a bed, a fridge, a stove, a couch, and a television. Where might he be? \\
      & \textbf{Answer options:} (A) motel (B) school (C) hotel (D) apartment (E) friend's house  \\
      & \textbf{Model prediction:} (D) apartment
    \end{tabular}
  \end{minipage}
} \\
\midrule
\multicolumn{2}{l}{
  \begin{minipage}{\dimexpr 15cm + 2\tabcolsep + \arrayrulewidth\relax}
    \begin{tabular}{@{} p{2cm} p{13cm} @{}}
      \textbf{Intervened} 
      & \textbf{Question:} There was only one \textcolor{blue}{\textbf{cozy}} room in the place where Bill slept. It had a bed, a fridge, a stove, a couch, and a television. Where might he be? \\
      & \textbf{Answer options:} (A) motel (B) school (C) hotel (D) apartment (E) friend's house  \\
      & \textbf{Model prediction:} \textcolor{blue}{\textbf{(A) motel}}
    \end{tabular}
  \end{minipage}
} \\
\midrule
\multicolumn{2}{l}{
  \begin{minipage}{\dimexpr 15cm + 2\tabcolsep + \arrayrulewidth\relax}
    \textbf{Initial Explanation:} Bill's room contains amenities typical of a motel, such as a bed, fridge, stove, couch, and television, which are not usually found together in a school setting. \textcolor{red}{\textbf{Unfaithful}}
  \end{minipage}
} \\
\midrule
\multicolumn{1}{c}{\textbf{Natural Language Feedback (NLF)}} & \multicolumn{1}{c}{\textbf{Important-Word Feedback (IWF-Attn)}} \\
\midrule
\textbf{Feedback 1:} The explanation accurately reflects the reasoning behind selecting option (A) as it highlights the motel-like amenities present in Bill's room. However, it could be improved by explicitly mentioning that the combination of a bed, fridge, stove, couch, and television is more characteristic of a motel or hotel room than other options like a school or apartment. This would make the reasoning clearer and more convincing. No other significant improvements are needed. 
& \textbf{Feedback 1:} The 5 most important words that contributed to your prediction are: one, a, cozy, be, there. \\
\midrule
\textbf{Refinement 1:} Bill's room contains a bed, fridge, stove, couch, and television, which are typical amenities found in a motel or hotel room, not in a school or apartment setting. This combination of features makes option (A) Motel the most appropriate choice. \textcolor{red}{\textbf{Unfaithful}}
& \textbf{Refinement 1:} Bill's room, described as \textcolor{blue}{\textbf{cozy}} with amenities like a bed, fridge, stove, couch, and television, suggests a motel setting, as these features are commonly found together in such accommodations. \textcolor{green}{\textbf{Faithful}} \\
\midrule
\multicolumn{1}{c}{\textbf{......}}
& \multicolumn{1}{c}{\textbf{......}} \\
\midrule
\textbf{Refinement after 3 rounds:} Bill's room, equipped with a bed, fridge, stove, couch, and television, is indicative of a motel or hotel room, where such amenities are common for temporary accommodation and comfort. The presence of a bed, fridge, and stove suggests a living space, which aligns more with a motel setting than a school or apartment. This combination of features makes option (A) Motel the most appropriate choice. \textcolor{red}{\textbf{Unfaithful}} 
& \textbf{Refinement after 3 rounds:} Bill's \textcolor{blue}{\textbf{cozy}} room, complete with a bed, fridge, stove, couch, and television, suggests a motel setting, where such a combination of amenities is commonly found. \textcolor{green}{\textbf{Faithful}} \\
\bottomrule
\end{tabular}
\caption{Case study comparing NLF and IWF-Attn (our best-performing variant) on the ECQA dataset. The intervened word, highlighted in \textcolor{blue}{\textbf{blue}}, successfully changes the model prediction. \textcolor{green}{\textbf{Faithful}} indicates the explanation/refinement is faithful as judged by the counterfactual test, while \textcolor{red}{\textbf{Unfaithful}} indicates the opposite.}
\label{tab:case_study}
\end{table*}

\paragraph{Case Study}
Table~\ref{tab:case_study} presents a case study from the ECQA dataset comparing NLF and IWF-Attn. Starting from the same unfaithful initial explanation, IWF-Attn successfully achieves faithfulness after one refinement round guided by its identification of the five most important words for the prediction. In contrast, NLF fails to achieve faithfulness even after three rounds of refinement. Despite receiving detailed feedback suggesting various improvements, NLF's refinements become progressively longer but still remain unfaithful. Complete refinement details and additional examples from e-SNLI and ComVE can be found in Appendix~\ref{appendix:case_studies}.\looseness=-1

\section{Conclusion}
\label{sec:conclusion}
In this work, we presented SR-NLE, a framework for improving the faithfulness of NLEs through an iterative self-critique and refinement process. By enabling LLMs to iteratively refine their own explanations with self-feedback, our approach significantly reduces unfaithfulness rates across multiple datasets and models without requiring external supervision, additional training or architectural changes. Our experiments demonstrate that IWF consistently outperforms NLF, with attention-based methods achieving the best results. The detailed analysis reveals that our framework efficiently targets critical reasoning components, successfully converts unfaithful explanations to faithful ones, and optimizes explanation content rather than merely increasing length. These findings suggest that self-refinement offers a promising path toward more faithful explanation generation. Future work could explore additional feedback mechanisms and investigate the applicability of SR-NLE to a broader set of domains and diverse reasoning tasks.\looseness=-1

\section*{Limitations}
\label{sec:limitations}

While our SR-NLE framework shows promising improvements in explanation faithfulness, it has several limitations. 

\paragraph{Explanation Paradigm}
Our experiments focus only on post-hoc natural language explanations, where explanations are generated after prediction. It remains unclear whether our refinement process generalizes to other explanation paradigms, such as jointly generated rationales or chain-of-thought reasoning. Different explanation generation strategies might present unique challenges and opportunities for refinement that are not addressed in our current framework.

\paragraph{Evaluation Method}
We rely on counterfactual tests as the sole evaluation method for measuring explanation faithfulness. While this metric offers objective signals aligned with our goal, it reflects only one type of faithfulness criterion. Future work could explore additional automatic tests--such as consistency tests and simulatability tests--to provide a more comprehensive view of explanation faithfulness. Moreover, our evaluation approach does not capture other important aspects of explanation quality, such as plausibility, completeness, or alignment with human-annotated references.

\paragraph{Attribution Method}  
The effectiveness of our attribution-based IWF methods depends on the reliability of the underlying attribution techniques. Attention weights may not consistently reflect true feature importance, while integrated gradients can be sensitive to baseline choices and implementation details. In practice, applying integrated gradients to large language models often requires a substantial number of integration steps to achieve convergence, which increases computational cost and may limit scalability. As a result, the quality and efficiency of the feedback depend on how accurately these methods capture the model's actual reasoning process.

\paragraph{Model Scale and Architecture}
All of our experiments are conducted on LLMs in the 10B parameter range. Further investigation is needed to understand how model scale affects both the baseline quality of explanations and the effectiveness of self-refinement, especially for smaller open-weight models. The performance of SR-NLE might vary significantly with larger, more advanced models or different architectural designs.

\bibliography{custom}

\appendix
\section*{Appendix}
\section{Datasets}
\label{appendix:datasets}

\paragraph{Dataset Selection.}
For our experiments, we selected the first 1,000 instances from each dataset's test set. The full test set of ComVE contains 1,000 instances, while ECQA and e-SNLI have 2,194 and 9,824 instances, respectively. To verify the representativeness of these subsets, we analyzed their label distributions compared to the full test sets, as shown in Table~\ref{tab:dataset_analysis}. The slight deviation from a 50/50 split in ComVE results from the random ordering process during dataset preparation, where we randomly positioned the sentence that violates common sense as either the first or second sentence.

\begin{table}[ht]
\centering
\renewcommand{\arraystretch}{1.0}
\setlength{\tabcolsep}{6pt}
\begin{tabular}{llcc}
\toprule
\textbf{Dataset} & \textbf{Label} & \textbf{Subset} & \textbf{Full} \\
\midrule
\multirow{2}{*}{ComVE} 
 & Sentence 0 & 48.0 & 48.0 \\
 & Sentence 1 & 52.0 & 52.0 \\
\midrule
\multirow{5}{*}{ECQA} 
 & Option A & 20.5 & 20.8 \\
 & Option B & 22.5 & 19.6 \\
 & Option C & 16.4 & 18.6 \\
 & Option D & 22.3 & 21.5 \\
 & Option E & 18.3 & 19.6 \\
\midrule
\multirow{3}{*}{e-SNLI} 
 & Contradiction & 34.4 & 34.3 \\
 & Neutral & 32.7 & 32.8 \\
 & Entailment & 32.9 & 32.9 \\
\bottomrule
\end{tabular}
\caption{Label distribution comparison (\%) between our experimental subsets and full test sets.}
\label{tab:dataset_analysis}
\end{table}

\section{Models}
\label{appendix:models}

\paragraph{Model Specifications.}
Table~\ref{tab:model_versions} presents the specific versions and parameter sizes of the instruction-tuned models used in our experiments. All models were accessed through their Hugging Face\footnote{\url{https://huggingface.co}} implementations.

\paragraph{SentenceBERT Model.}
We use the SentenceBERT model \textbf{all-mpnet-base-v2} as the semantic encoder for the centroid voting method in the SC-NLE baseline. This model was selected based on its strong performance in various semantic similarity tasks.

\begin{table}[t]
\centering
\renewcommand{\arraystretch}{1.0}
\setlength{\tabcolsep}{8pt}
\begin{tabular}{llc}
\toprule
\textbf{Model} & \textbf{Version} & \textbf{Size}\\
\midrule
Falcon & Falcon3-Instruct & 7B\\
Llama & Llama3.1-Instruct & 8B\\
Mistral & Mistral-Instruct-v0.3 & 7B\\
Qwen & Qwen2.5-Instruct & 7B\\
\bottomrule
\end{tabular}
\caption{Details of the models used in our experiments.}
\label{tab:model_versions}
\end{table}

\section{Evaluation}
\label{appendix:evaluation}

This section details our process for generating interventions for the counterfactual test used in our faithfulness evaluation.

\paragraph{Prompting Strategy.}
We used GPT-4o \cite{openai2024gpt4technicalreport} to generate interventions by adding adjectives before nouns or adverbs before verbs. For datasets with paired input texts (ComVE and e-SNLI), we generated 10 interventions for each text (e.g., 10 for premise and 10 for hypothesis in e-SNLI), resulting in 20 total interventions per instance. For ECQA, which has a single input text, we generated all 20 interventions for the same text. The prompt template used for intervention generation is shown in Table~\ref{tab:intervention_prompt}.

\paragraph{Intervention Quality Analysis.} 
To verify the quality of our generated interventions, we manually examined a random sample of 50 instances from each dataset (150 total). Our analysis confirmed that the intervened instances remained meaningful and coherent, with exactly one word modified as intended. These quality checks ensured that our interventions were suitable for faithfulness evaluation, as they created meaningful variations that could potentially change model predictions.

\begin{table}[t]
\centering
\small
\begin{tabular}{@{}%
>{\raggedright\arraybackslash}p{\linewidth}%
@{}}
\toprule
\textbf{Task:} \\
You will be given a sentence. Your task is to edit the sentence by inserting a random adjective before a noun or a random adverb before a verb. The noun or verb must be selected randomly from the given sentence. \\
\\
\textbf{Requirements:} \\
- Generate 10 different edits. \\
- Each edit should modify only one word. \\
- Enclose only the modified word in square brackets [ ]. \\
- Ensure that the sentence remains grammatically correct and natural. \\
\\
\textbf{Output format:} \\
1. [Edited Sentence] \\
2. [Edited Sentence] \\
... \\
10. [Edited Sentence] \\
\\
\textbf{Sentence:} \\
\{sentence to be edited\} \\
\bottomrule
\end{tabular}
\caption{Prompt template for generating interventions. For ECQA, we modified the prompt to request 20 edits instead of 10.}
\label{tab:intervention_prompt}
\end{table}

\section{Attribution Methods}
\label{appendix:attribution}

\paragraph{Integrated Gradients}
We implement Integrated Gradients (IG) following \citet{sundararajan2017axiomaticattributiondeepnetworks} to compute token importance. Since all our models are generative language models, we use the end-of-sequence (EOS) token embedding as the baseline, as it serves as a neutral and consistently defined default signal.

\paragraph{Attention} 
We leverage the model's self-attention mechanism to measure token importance. Specifically, we extract attention weights from the final layer of the model and average them across all attention heads. For each target token, these weights indicate how much the model attended to each input token when generating that token.

\section{Ablation Studies}
\label{appendix:ablations}

All ablation studies are conducted on the same 100-instance subsets sampled from each dataset. For the IG integration steps (Section~\ref{appendix:ablation_ig_steps}), we directly use these 100 original instances. For the other two ablation studies (Section~\ref{appendix:ablation_sc} and~\ref{appendix:ablation_topn}), we follow the same procedure as in the main experiments to compute unfaithfulness: for each of the 100 instances, we generate 20 interventions (totalling 2000 intervened instances), and then select and perform experiments on the counter instances. It's important to note that we conducted these ablation studies across all dataset and model combinations. Therefore, all metrics reported in these sections represent averaged values across the entire experimental matrix, providing a comprehensive view of our method's performance across different conditions.

\subsection{Integration Steps for IG Attribution}
\label{appendix:ablation_ig_steps}

\begin{figure}[t]
\centering
\includegraphics[width=\linewidth]{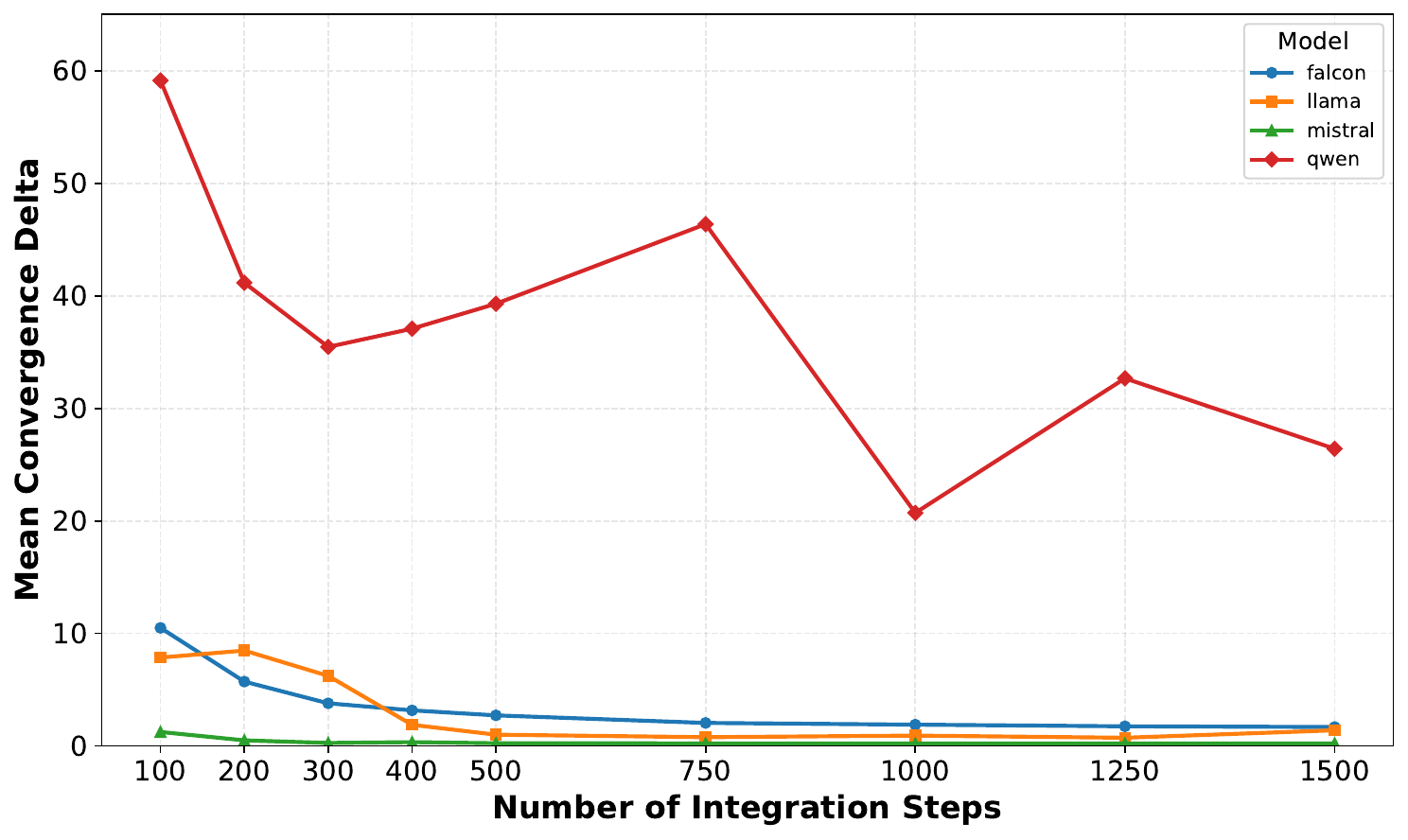}
\caption{Impact of number of integration steps on convergence of IG attribution methods. Convergence delta measures the approximation error in the numerical integration. Lower values indicate more stable and accurate attribution calculations.}
\label{fig:ig_steps}
\end{figure}

We investigated the impact of integration steps on the convergence of IG attribution calculations by experimenting with nine different step settings ranging from 100 to 1500. Figure \ref{fig:ig_steps} shows that most models (Falcon, Llama, and Mistral) reach reasonable convergence around 500 integration steps, with the convergence delta showing minimal changes beyond this point. In contrast, the Qwen model exhibits higher variability and slower convergence, requiring more steps to achieve stable attribution values. This may be caused by architectural differences that affect how gradients are calculated and propagated through the model. Based on these observations, we selected 500 integration steps for Falcon, Llama, and Mistral, while using 1000 steps for Qwen in our main experiments.

\subsection{SC-NLE Parameters}
\label{appendix:ablation_sc}

\begin{figure}[t]
  \centering
  \includegraphics[width=\linewidth]{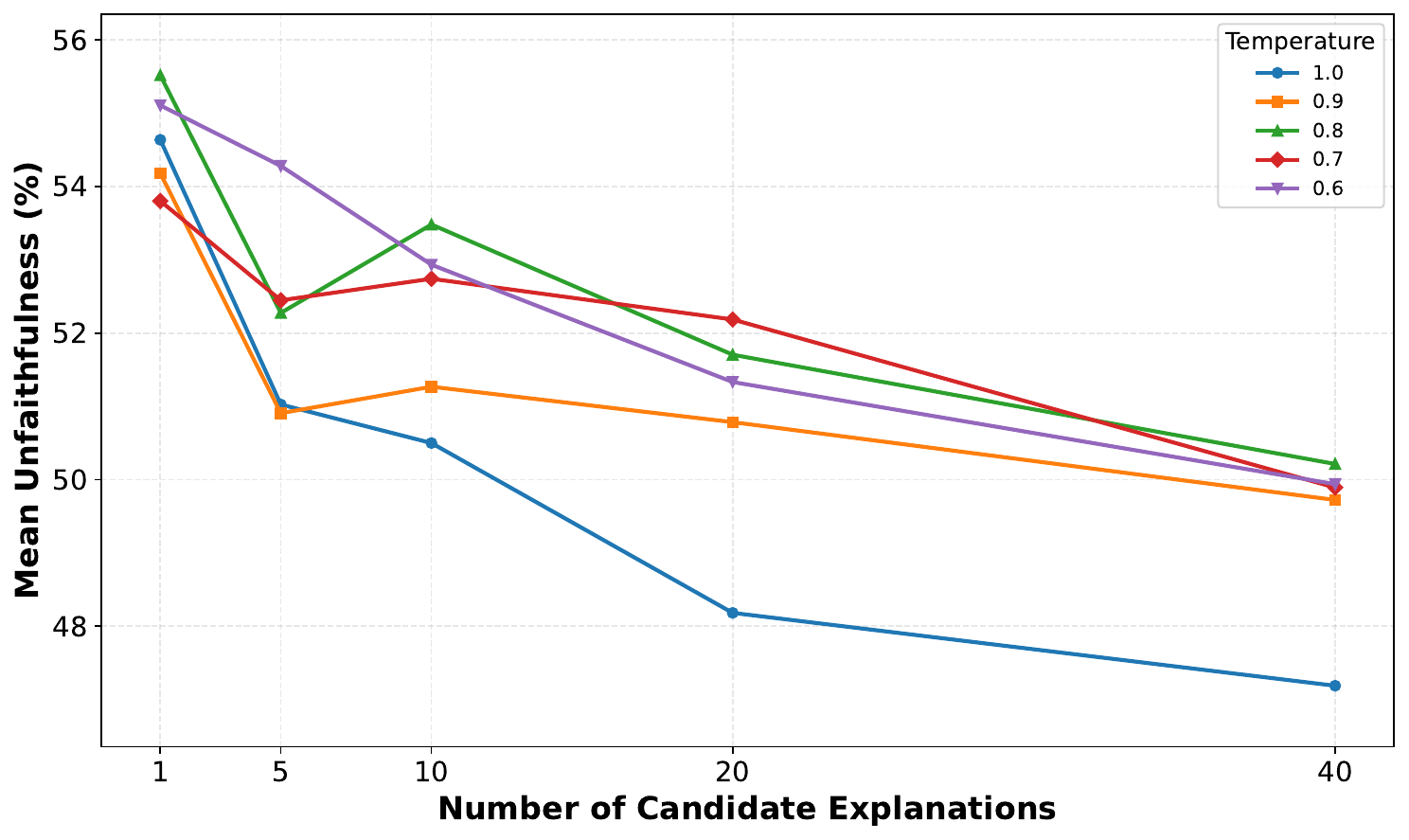}
  \caption{Impact of sampling temperature and candidate count on SC-NLE unfaithfulness.}
  \label{fig:sc_params}
\end{figure}

We investigated the impact of two key parameters for our SC-NLE baseline: candidate explanation count and sampling temperature. Figure \ref{fig:sc_params} shows that increasing the number of candidates reduces unfaithfulness, with significant improvements up to 20 samples. Temperature also affects performance, with temperature 1.0 consistently outperforming lower values, especially at higher sample counts. Based on these results, we selected 20 candidate explanations at temperature 1.0, balancing performance improvements with computational cost.

\subsection{Number of Important Words}
\label{appendix:ablation_topn}

\begin{figure*}[t]
  \centering
  \begin{minipage}{0.48\linewidth}
    \centering
    \includegraphics[width=\linewidth]{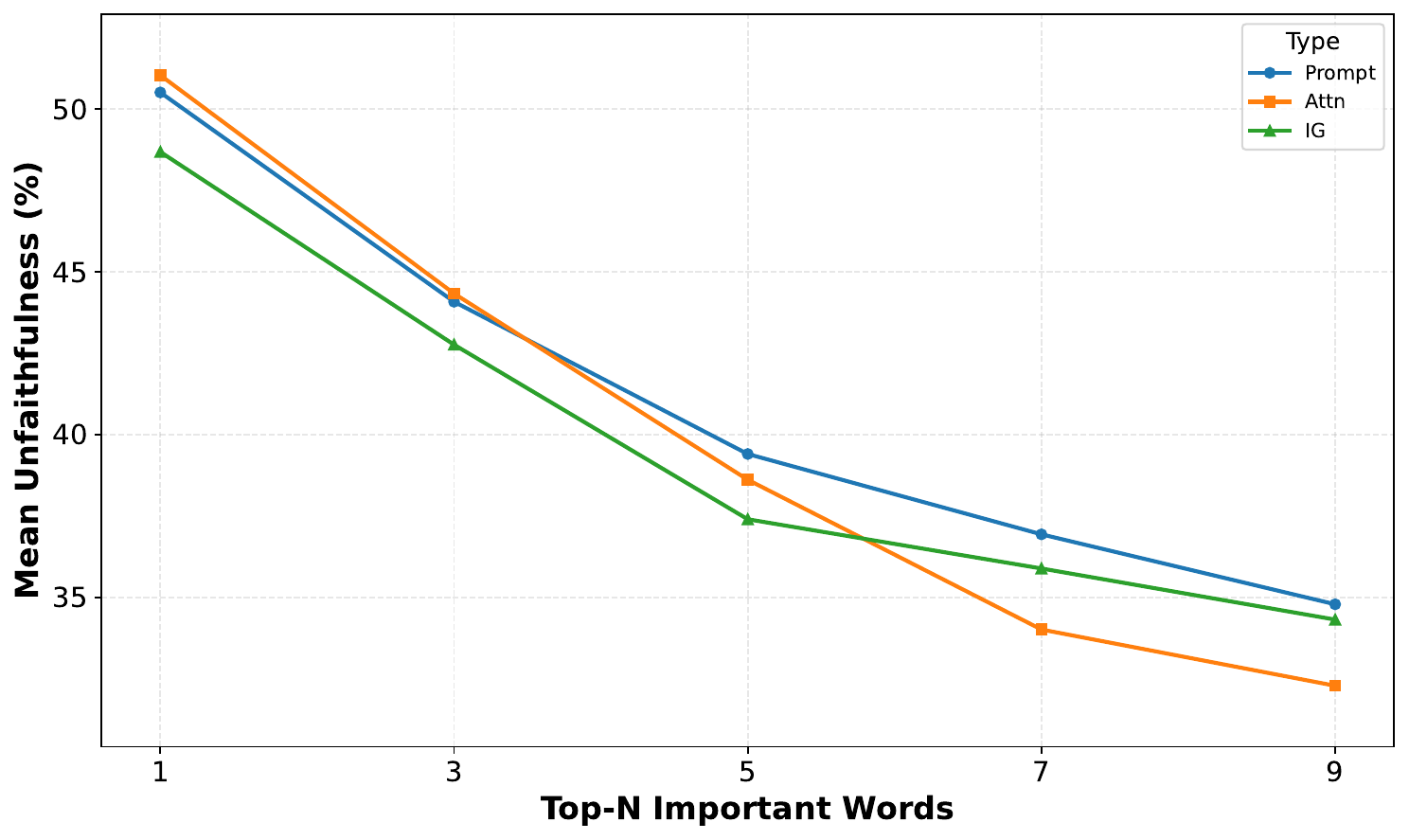}
    \vspace{-1mm}
    \small (a) Mean Unfaithfulness vs. Top-N.
  \end{minipage}
  \hfill
  \begin{minipage}{0.48\linewidth}
    \centering
    \includegraphics[width=\linewidth]{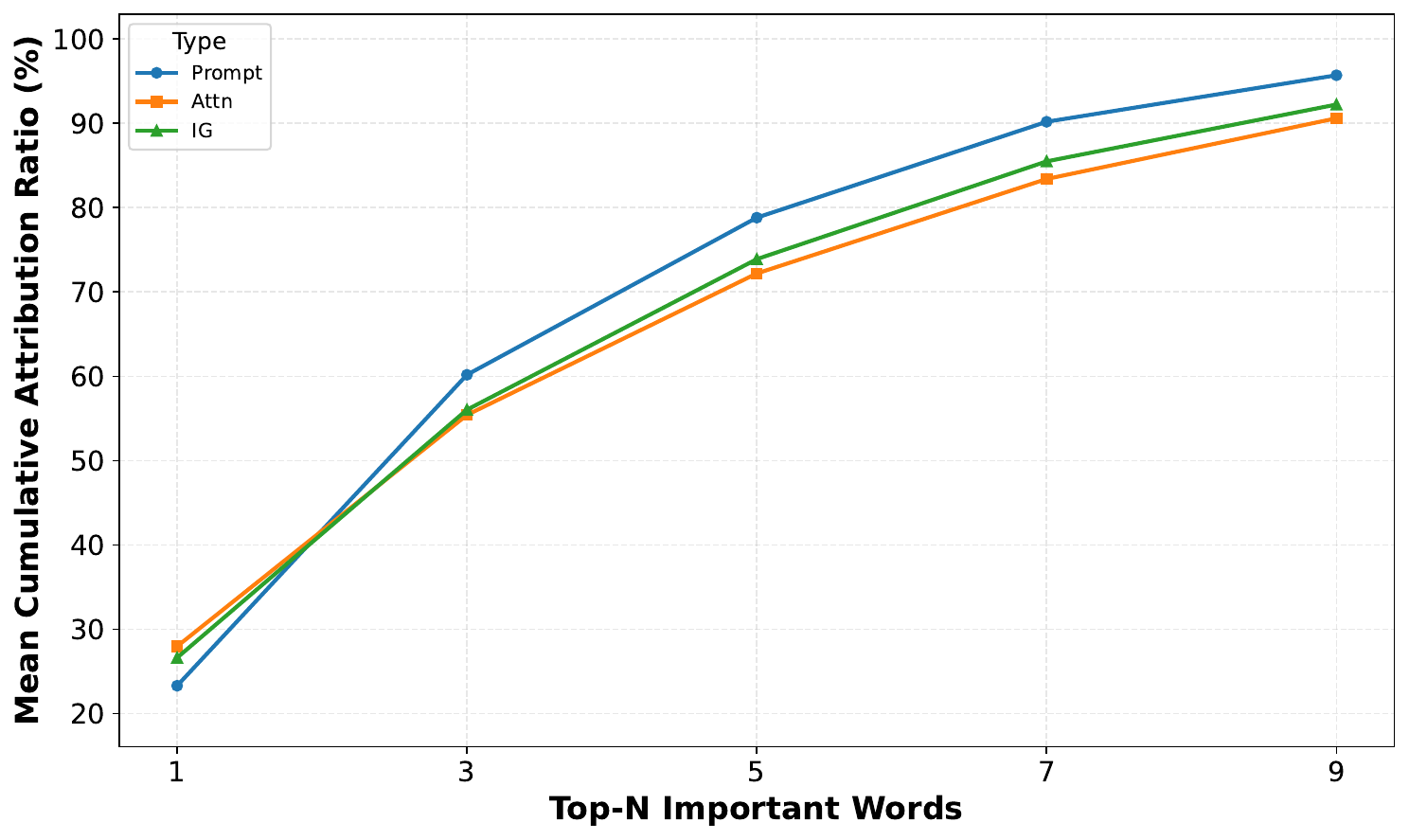}
    \vspace{-1mm}
    \small (b) Mean Cumulative Attribution Ratio vs. Top-N.
  \end{minipage}
  \vspace{-1mm}
  \caption{Analysis of top-N important words selection. (a) Shows how unfaithfulness decreases with increasing N across feedback types (lower is better). (b) Depicts the proportion of total attribution captured by top-N words across attribution methods (higher indicates greater coverage).}
  \label{fig:topn}
\end{figure*}

We investigated the optimal number of important words (top-$N$) for our Important Word Feedback through two complementary analyses: examining unfaithfulness changes and analyzing attribution distribution patterns. Figure (\ref{fig:topn}a) shows that unfaithfulness decreases as $N$ increases from 1 to 9 across all feedback types, with the most significant improvements occurring between $N=1$ and $N=5$. Concurrently, Figure (\ref{fig:topn}b) reveals that the top-5 important words capture approximately 70-80\% of the total attribution magnitude, despite typically representing only a small fraction of input tokens. Therefore, we selected $N=5$ for all our main experiments.

\section{Additional Results}
\label{appendix:results}

\subsection{Prediction Accuracy}
Table~\ref{tab:prediction_accuracy} shows the prediction accuracy for each model across the three datasets. While accuracy is not directly related to our evaluation focus, we can observe that models demonstrate strong performance. Across different datasets, Falcon and Qwen models generally achieve higher accuracy than Llama and Mistral models. Notably, on the ComVE dataset, all models achieve accuracy rates above 90\%.

\begin{table}[ht]
\centering
\renewcommand{\arraystretch}{1.0}
\setlength{\tabcolsep}{6pt}
\begin{tabular}{lcccc}
\toprule
& \textbf{Falcon} & \textbf{Llama} & \textbf{Mistral} & \textbf{Qwen} \\
\midrule
\textbf{ComVE}  & 96.70 & 90.80 & 94.50 & 96.70 \\
\textbf{ECQA}   & 77.10 & 73.20 & 68.34 & 79.50 \\
\textbf{e-SNLI} & 89.60 & 56.90 & 58.10 & 88.70 \\
\bottomrule
\end{tabular}
\caption{Model prediction accuracy (\%).}
\label{tab:prediction_accuracy}
\end{table}

\subsection{Counter Rates}
Table~\ref{tab:counter_rate} shows the number of counter instances and counter rates for each model-dataset combination out of 20,000 total intervened instances. Each model achieves counter rates of 10-15\% on the ECQA and e-SNLI datasets, while on ComVE, the rates are generally below 10\%, with most under 5\%.

\begin{table}[ht]
\centering
\renewcommand{\arraystretch}{1.0}
\setlength{\tabcolsep}{6pt}
\begin{tabular}{lcccc}
\toprule
& \textbf{Falcon} & \textbf{Llama} & \textbf{Mistral} & \textbf{Qwen} \\
\midrule
\multirow{2}{*}{\textbf{ComVE}} & 392 & 1244 & 829 & 456 \\[-0.3em]
                              & \multicolumn{1}{r}{\footnotesize (1.96)} & \multicolumn{1}{r}{\footnotesize (6.22)} & 
                              \multicolumn{1}{r}{\footnotesize (4.15)} & 
                              \multicolumn{1}{r}{\footnotesize (2.28)} \\
\midrule
\multirow{2}{*}{\textbf{ECQA}} & 2305 & 2418 & 2377 & 2150 \\[-0.3em]
                             & \multicolumn{1}{r}{\footnotesize (11.53)} & \multicolumn{1}{r}{\footnotesize (12.09)} & 
                             \multicolumn{1}{r}{\footnotesize (11.92)} & 
                             \multicolumn{1}{r}{\footnotesize (10.75)} \\
\midrule
\multirow{2}{*}{\textbf{e-SNLI}} & 2812 & 2298 & 2476 & 3058 \\[-0.3em]
                               & \multicolumn{1}{r}{\footnotesize (14.06)} & \multicolumn{1}{r}{\footnotesize (11.49)} & \multicolumn{1}{r}{\footnotesize (12.38)} & \multicolumn{1}{r}{\footnotesize (15.29)} \\
\bottomrule
\end{tabular}
\caption{Number of counter instances (top) and counter rates in \% (bottom, in parentheses) for each model-dataset combination out of 20,000 total instances.}
\label{tab:counter_rate}
\end{table}

\subsection{Average Sequence Lengths}
Table~\ref{tab:seq_len} reports the average sequence lengths of counter instances for each model-dataset combination. We provide results under two settings: \textbf{\textit{Full}}, which denotes the total word length of the input, and \textbf{\textit{Unique}}, which denotes the word length after removing duplicate words. The lengths range from about 13 to 21 words under the \textbf{\textit{Full}} setting, while the \textbf{\textit{Unique}} setting is consistently shorter, around 9 to 16 words.

\begin{table}[ht]
\centering
\renewcommand{\arraystretch}{1.0}
\setlength{\tabcolsep}{6pt}
\begin{tabular}{lcccc}
\toprule
& \textbf{Falcon} & \textbf{Llama} & \textbf{Mistral} & \textbf{Qwen} \\
\midrule
\multicolumn{5}{c}{\textbf{\textit{Full}}} \\
\midrule
\textbf{ComVE}  & 15.72 & 15.13 & 14.96 & 16.18 \\
\textbf{ECQA}   & 12.99 & 13.53 & 13.45 & 13.64 \\
\textbf{e-SNLI} & 21.08 & 21.55 & 20.70 & 21.35 \\
\midrule
\multicolumn{5}{c}{\textbf{\textit{Unique}}} \\
\midrule
\textbf{ComVE}  & 10.16 & 9.35 & 9.67 & 10.26 \\
\textbf{ECQA}  & 12.25 & 12.66 & 12.55 & 12.82 \\
\textbf{e-SNLI} & 15.65 & 16.28 & 15.78 & 15.81 \\
\bottomrule
\end{tabular}
\caption{Average sequence lengths of counter instances. \textbf{\textit{Full}} refers to the total word length, while \textbf{\textit{Unique}} refers to the word length after removing duplicate words.}
\label{tab:seq_len}
\end{table}

\subsection{Intermediate Results}
The unfaithfulness rates after refinement round 1 and refinement round 2 are shown in Table~\ref{tab:round_results}. We can observe a clear downward trend in unfaithfulness rates across successive refinement rounds, demonstrating the progressive effectiveness of our iterative approach.

\begin{table*}[ht]
\centering
\renewcommand{\arraystretch}{1.4}
\setlength{\tabcolsep}{2.1pt}
\small
\begin{tabular*}{\linewidth}{@{\extracolsep{\fill}}l@{\hspace{4pt}}l|cccc|cccc|cccc|c@{}}
\toprule
\multicolumn{2}{@{}c|}{\multirow{2}{*}{\textbf{Method}}} & \multicolumn{4}{c|}{\textbf{ComVE}} & \multicolumn{4}{c|}{\textbf{ECQA}} & \multicolumn{4}{c|}{\textbf{e-SNLI}} & \multirow{2}{*}{\textbf{Avg.}} \\
\cmidrule{3-14}
\multicolumn{2}{@{}c|}{} & Falcon & Llama & Mistral & Qwen & Falcon & Llama & Mistral & Qwen & Falcon & Llama & Mistral & Qwen & \\

\midrule
\multicolumn{15}{c}{\textbf{R1}} \\
\midrule
& NLF        & 65.05 & 67.60 & 67.91 & 63.60   & 46.03 & 39.83 & 45.90 & 49.72    & 22.65 & 51.96 & 48.42 & 38.98 & 50.64 \\
& IWF-Pmt    & \underline{51.28} & 63.67 & \underline{51.63} & \underline{56.58}   & \textbf{22.60} & \textbf{24.94} & 40.60 & 32.70    & 18.88 & \underline{38.51} & \underline{39.54} & 27.17 & 39.01 \\
& IWF-Attn   & 53.83 & \underline{62.38} & \textbf{50.42} & \textbf{54.39}   & 27.29 & \underline{25.19} & \textbf{40.34} & \textbf{29.44}    & \textbf{17.25} & \textbf{37.55} & 40.35 & \textbf{23.38} & \textbf{38.48} \\
& IWF-IG     & \textbf{48.98} & \textbf{61.17} & 54.16 & 57.46   & \underline{24.64} & 25.35 & \underline{40.47} & \underline{30.56}    & \underline{17.89} & 39.69 & \textbf{39.18} & \underline{26.03} & \underline{38.80} \\

\midrule
\multicolumn{15}{c}{\textbf{R2}} \\
\midrule
& NLF        & 61.99 & 64.95 & 65.38 & 59.65   & 44.38 & 38.50 & 45.69 & 47.86   & 22.97 & 48.87 & 45.92 & 36.72 & 48.57 \\
& IWF-Pmt    & \underline{46.17} & 62.70 & \underline{46.56} & \underline{53.95}   & \textbf{23.95} & \textbf{24.32} & \textbf{42.03} & 29.77   & 20.66 & \underline{36.34} & 38.45 & 24.75 & 37.47 \\
& IWF-Attn   & 50.26 & \underline{60.45} & \textbf{45.11} & \textbf{51.10}   & 27.25 & \underline{24.40} & 42.91 & \textbf{26.70}   & \textbf{17.14} & \textbf{35.16} & \underline{38.41} & \textbf{20.01} & \textbf{36.58} \\
& IWF-IG     & \textbf{43.88} & \textbf{58.68} & 49.58 & 54.82   & \underline{24.69} & 25.06 & \underline{42.32} & \underline{28.23}   & \underline{18.28} & 37.60 & \textbf{36.51} & \underline{22.47} & \underline{36.84} \\


\bottomrule
\end{tabular*}
\caption{Unfaithfulness rates (\%) after refinement rounds 1 (R1) and refinement round 2 (R2). Best (lowest) results per dataset-model combination are \textbf{bolded}, second best are \underline{underlined}.}
\label{tab:round_results}
\end{table*}

\subsection{Reliability of Prompt-based Important Words Selection Strategy}
\label{appendix:hallucination_rate}
To assess the reliability of prompt-based important words selection, we examined whether the important words identified by IWF-Pmt are grounded in the input. Specifically, we measured the hallucination rate, defined as the proportion of top-5 selected words that do not appear in the input. Table~\ref{tab:hallucination_rate} presents the results across all dataset-model combinations. While we observe some variation—with ECQA showing slightly higher rates and Qwen exhibiting more hallucination compared to other models—the overall average hallucination rate across the 12 combinations is only 3.75\%. This low rate demonstrates that prompt-based word selection is highly reliable, with the selected words being well-grounded in the input.

\begin{table}[ht]
\centering
\renewcommand{\arraystretch}{1.0}
\setlength{\tabcolsep}{6pt}
\begin{tabular}{lcccc}
\toprule
& \textbf{Falcon} & \textbf{Llama} & \textbf{Mistral} & \textbf{Qwen} \\
\midrule
\textbf{ComVE}  & 0.01 & 0.01 & 0.01 & 0.05 \\
\textbf{ECQA}   & 0.07 & 0.10 & 0.03 & 0.08 \\
\textbf{e-SNLI} & 0.01 & 0.01 & 0.02 & 0.05 \\
\bottomrule
\end{tabular}
\caption{Hallucination rates (\%) of extracted important words in IWF-Pmt, 
measured as the proportion of top-5 words not appearing in the input.}
\label{tab:hallucination_rate}
\end{table}

\subsection{Word Selection Quality Analysis}
\label{appendix:selection_quality}

To assess the impact of word selection quality on IWF performance, we conducted experiments comparing different selection methods against a random baseline. Specifically, we randomly selected five words from the input and applied three rounds of refinement on the e-SNLI dataset across all four models, averaging results over three random seeds. As shown in Table~\ref{tab:random}, while the random baseline shows slightly higher unfaithfulness rates, it achieves performance close to both prompt-based and attribution-based IWF methods. To understand this result, we analyzed how often the intervened word (i.e., the true reasoning factor that made the label change) appeared in the top-$N$ selected words. Table~\ref{tab:inclusion} shows that current selection methods—whether prompt-based or attribution-based—capture the true reasoning word at rates similar to random selection. These findings reveal two important insights:
\begin{enumerate}
    \item \textbf{Robustness of IWF}: Even with suboptimal word selection, IWF can effectively improve explanation faithfulness, demonstrating that the iterative refinement process in SR-NLE contributes significantly to performance improvements beyond the quality of word attribution.\looseness=-1
    
    \item \textbf{Opportunities for improvement}: Current word attribution methods have considerable room for enhancement. As better attribution techniques are developed, IWF could potentially achieve even stronger performance within our SR-NLE framework.
\end{enumerate}

\noindent This analysis underscores that IWF's effectiveness stems from the iterative refinement process in our SR-NLE framework rather than perfect word identification, making IWF a robust feedback mechanism that can benefit from future advances in attribution methods.

\begin{table}[t]
\centering
\renewcommand{\arraystretch}{1.3}
\setlength{\tabcolsep}{4pt}
\small
\begin{tabular}{lccccc}
\toprule
& \textbf{Falcon} & \textbf{Llama} & \textbf{Mistral} & \textbf{Qwen} & \textbf{Avg.} \\
\midrule
Random   & \makecell{21.18 \\ \hspace{0.8em}{\scriptsize$\pm$0.26}}
         & \makecell{35.32 \\ \hspace{0.8em}{\scriptsize$\pm$0.71}}
         & \makecell{37.68 \\ \hspace{0.8em}{\scriptsize$\pm$0.63}}
         & \makecell{22.91 \\ \hspace{0.8em}{\scriptsize$\pm$0.68}}
         & \makecell{29.28 \\ \hspace{0.8em}{\scriptsize$\pm$7.64}}
         \\
\midrule
IWF-Pmt  & 22.01 & 36.16 & 37.80 & 24.43 &  30.10\\
IWF-Attn & \textbf{18.21} & \textbf{34.94} & 38.49 & \textbf{19.10} &  \textbf{27.69}\\
IWF-IG   & 18.35 & 37.38 & \textbf{35.86} & 21.98 &  28.39\\
\bottomrule
\end{tabular}
\caption{Comparison of random baseline against IWF methods on e-SNLI dataset (unfaithfulness rates in \%). Random results show mean ± standard deviation over three seeds. All methods use three refinement rounds.}
\label{tab:random}
\end{table}

\begin{table}[t]
\centering
\renewcommand{\arraystretch}{1.2}
\setlength{\tabcolsep}{6pt}
\small
\begin{tabular}{lccccc}
\toprule
& \textbf{Top-1} & \textbf{Top-2} & \textbf{Top-3} & \textbf{Top-4} & \textbf{Top-5} \\
\midrule
Random & 6.58 & 13.28 & 20.12 & 27.03 & 34.00 \\
Pmt    & 3.25 & 8.70  & 14.89 & 22.17 & 30.51 \\
Attn   & 4.52 & 14.84 & \textbf{25.52} & \textbf{36.07} & \textbf{46.25} \\
IG     & \textbf{8.58} & \textbf{17.14} & 25.22 & 32.70 & 39.87 \\
\bottomrule
\end{tabular}
\caption{Intervened word inclusion rate (\%) in the top-$N$ selected words under the random baseline and IWF methods.}
\label{tab:inclusion}
\end{table}

\subsection{Performance Visualization}
\label{appendix:per_vis}

\begin{figure}[t]
    \centering
    \begin{subfigure}[b]{\linewidth}
        \centering
        \includegraphics[width=0.9\linewidth]{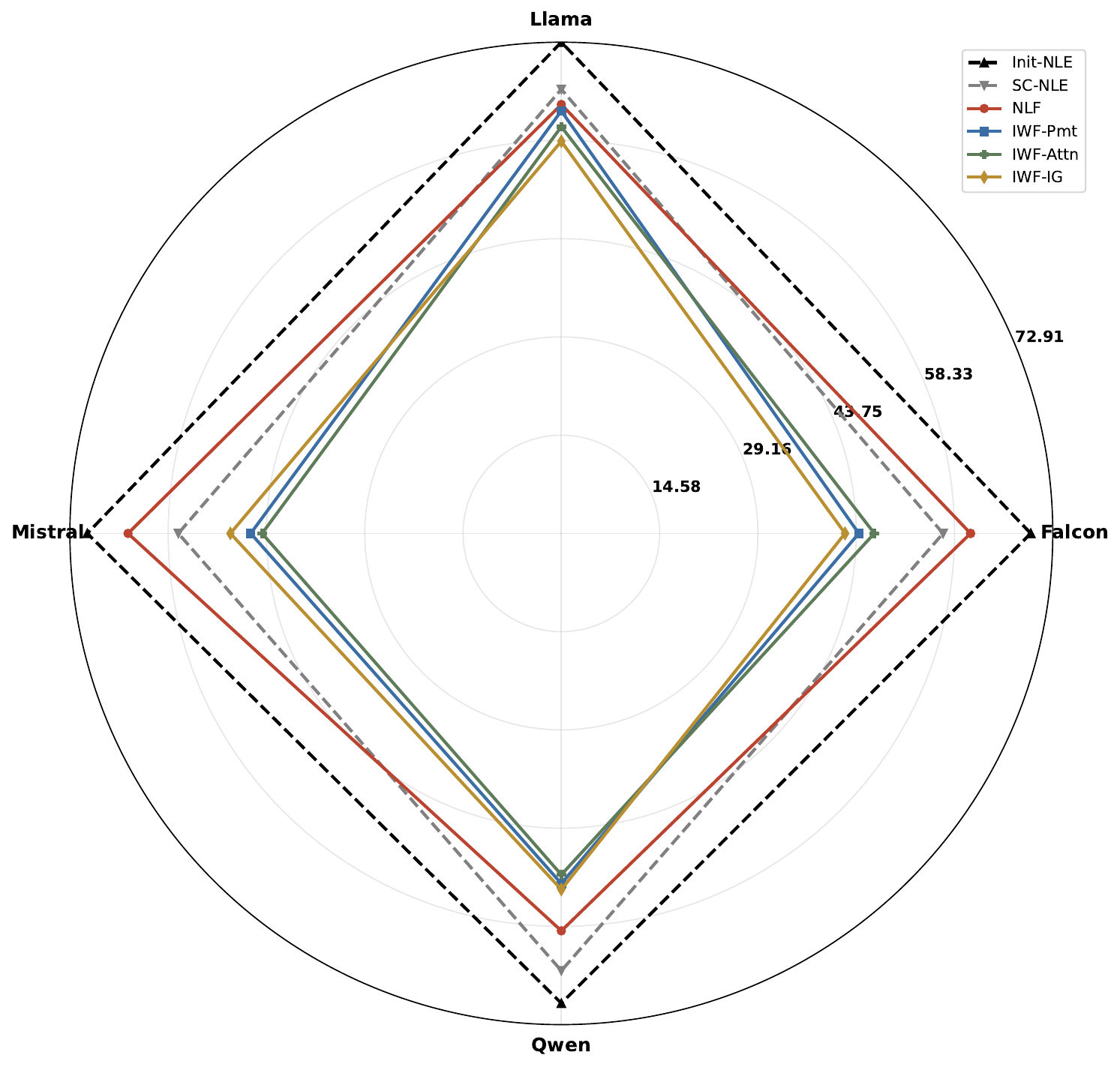}
        \caption{ComVE}
        \label{fig:radar_comve}
    \end{subfigure}

    \vspace{0.6em}

    \begin{subfigure}[b]{\linewidth}
        \centering
        \includegraphics[width=0.9\linewidth]{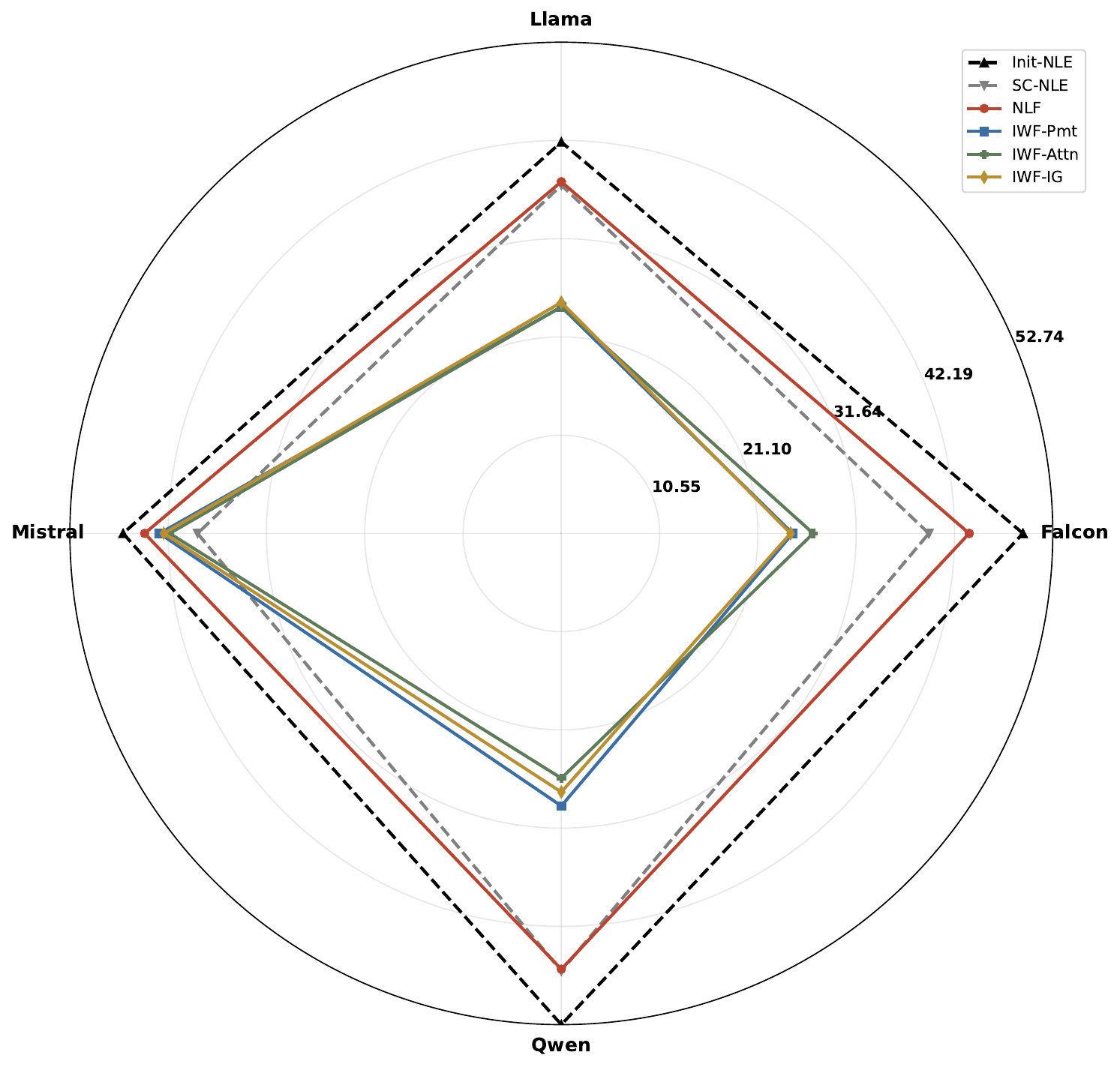}
        \caption{ECQA}
        \label{fig:radar_ecqa}
    \end{subfigure}

    \vspace{0.6em}

    \begin{subfigure}[b]{\linewidth}
        \centering
        \includegraphics[width=0.9\linewidth]{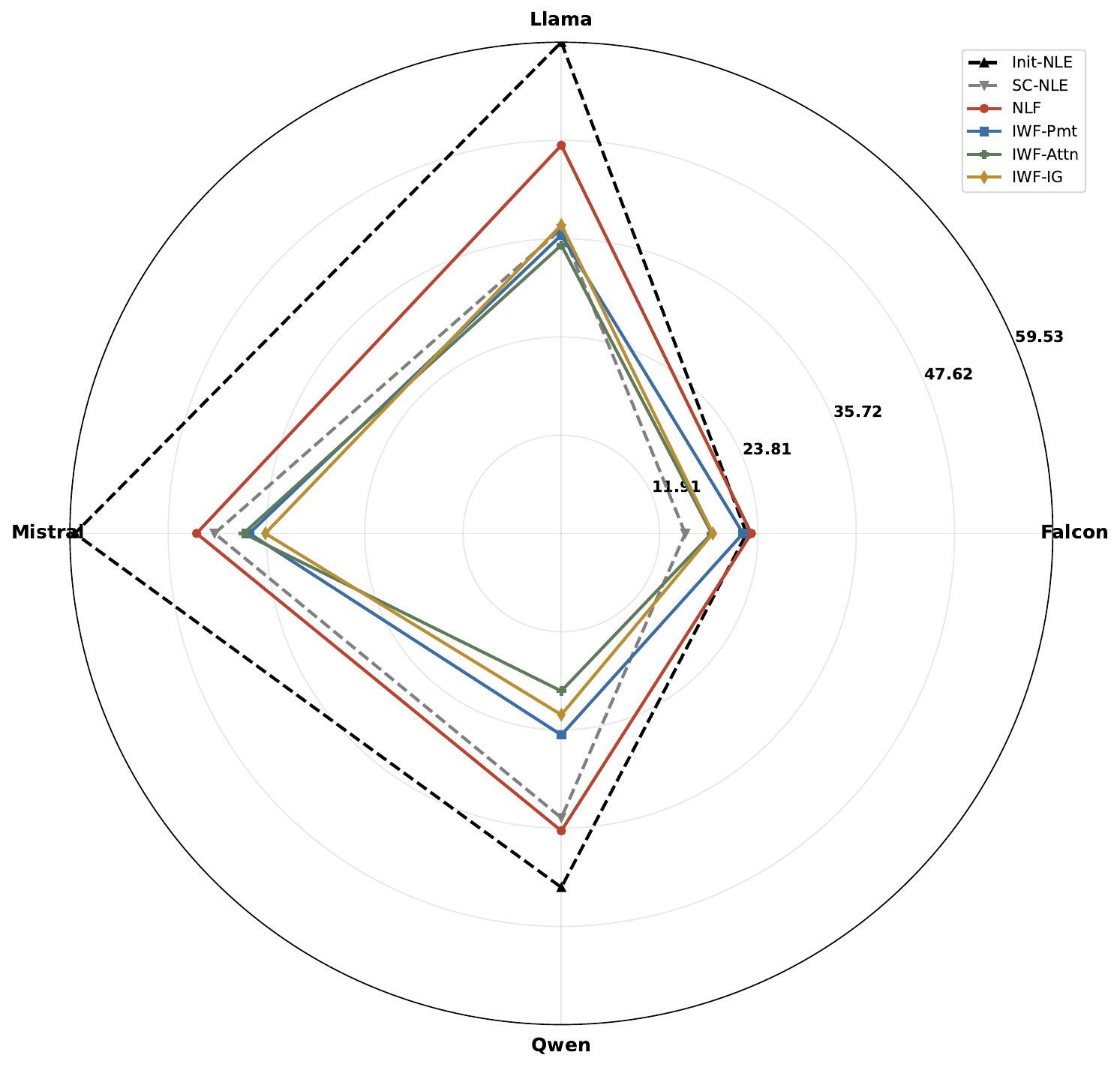}
        \caption{e-SNLI}
        \label{fig:radar_esnli}
    \end{subfigure}

    \caption{Radar chart visualization of unfaithfulness rates after 3 rounds of refinement. Lower values and smaller areas indicate better performance.}
    \label{fig:radar_charts}
\end{figure}

Figure~\ref{fig:radar_charts} presents radar chart visualizations of unfaithfulness rates after 3 rounds of refinement. In each chart, the four axes represent different models, while the connected areas represent different methods. The visualizations clearly show that our SR-NLE methods achieve smaller areas than the baselines in the majority of cases, aligning with our quantitative results and demonstrating the effectiveness of the framework across different datasets and models.\looseness=-1

\section{Algorithms}
\label{appendix:algorithms}

Algorithms~\ref{alg:srnle}-\ref{alg:semantic_centroid_voting} present the SR-NLE framework, attribution-based IWF \textsc{Score}, and semantic centroid voting strategy for SC-NLE, respectively.

\begin{algorithm*}[t]
\caption{SR-NLE Framework}
\label{alg:srnle}
\begin{algorithmic}[1]
\Require Model $\mathcal{M}$, prompts $\{p_{\text{ans}}, p_{\text{exp}}, p_{\text{fb}}, p_{\text{ref}}\}$, input $x$, feedback type $t \in \{\text{NLF}, \text{IWF}\}$, IWF method $m \in \{\text{prompt-based}, \text{attribution-based}\}$, refinement rounds $K$, number of important words $N$
\Ensure Final answer $y$ and refined explanation $e^K$
\State $y \leftarrow \mathcal{M}(p_{\text{ans}} \oplus x)$
\State $e^0 \leftarrow \mathcal{M}(p_{\text{exp}} \oplus x \oplus y)$
\If{$t = \text{IWF}$} \Comment{Prepare Important-Word Feedback following Eq.~\ref{eq:fb_iw}}
    \If{$m = \text{prompt-based}$}
        \State $\mathcal{S} \leftarrow \mathcal{M}(p_{\text{fb}} \oplus x \oplus y)$
    \Else
        \State $\mathcal{S} \leftarrow \textsc{AttributionScore}(\mathcal{M}, p_{\text{ans}}, x, y)$ \Comment{Algorithm \ref{alg:attr_iwf}}
    \EndIf
    \State $\mathcal{I} \leftarrow \textsc{Select}(\mathcal{S}, N)$
    \State $f_{\text{iw}} \leftarrow \textsc{Format}(\mathcal{I})$
\EndIf
\For{$r = 1$ to $K$} \Comment{Iterative refinement}
    \If{$t = \text{NLF}$}
        \State $f^r \leftarrow \mathcal{M}(p_{\text{fb}} \oplus x \oplus y \oplus e^{r-1})$
    \Else
        \State $f^r \leftarrow f_{\text{iw}}$
    \EndIf
    \State $e^r \leftarrow \mathcal{M}(p_{\text{ref}} \oplus x \oplus y \oplus e^{r-1} \oplus f^r)$
\EndFor
\State \Return $y, e^K$
\end{algorithmic}
\end{algorithm*}
\begin{algorithm*}[t]
\caption{Attribution-based IWF \textsc{Score}}
\label{alg:attr_iwf}
\begin{algorithmic}[1]
\Require Model $\mathcal{M}$, prompt $p_{\text{ans}}$, input $x$, answer $y$, attribution method $m \in \{\text{IG}, \text{Attention}\}$
\Ensure Word importance scores $\mathcal{S}$

\State Locate answer span $y$ within model output
\For{each token $y_j$ in answer span}
\Comment{Sequential target token attribution}
    \For{each token $x_i$ in full model input}
    \Comment{Token-level computation}
        \State $a_{i,j} \leftarrow |\text{Attribution}(x_i, y_j | \text{context}_{<j})|$ 
    \EndFor
\EndFor
\For{each token $x_i$ in full model input}
    \State $a_i \leftarrow \sum_{j=1}^{|y|} a_{i,j}$ \Comment{Target-level aggregation}
\EndFor
\For{each word $w$ in task input $x$}
    \State $\text{S}(w) \leftarrow \sum_{i \in \text{indices}(w)} a_i$ \Comment{Word-level aggregation}
\EndFor
\State \Return $\mathcal{S}$
\end{algorithmic}
\end{algorithm*}
\begin{algorithm*}[t]
\caption{Semantic Centroid Voting for SC-NLE}
\label{alg:semantic_centroid_voting}
\begin{algorithmic}[1]
\Require Candidate explanations $E = \{e_1, e_2, ..., e_n\}$, SentenceBERT model $M$
\Ensure Most representative explanation $e_{best}$
\State $embeddings \leftarrow M.encode(E)$ \Comment{Encode all explanations}
\State $centroid \leftarrow \frac{1}{n}\sum_{i=1}^{n} embeddings_i$ \Comment{Compute centroid embedding}
\State $similarities \leftarrow [\,]$ 
\For{$i = 1$ to $n$}
    \State $similarities_i \leftarrow \text{cosine\_similarity}(centroid, embeddings_i)$ \Comment{Compute similarity to centroid}
\EndFor
\State $best\_idx \leftarrow \arg\max(similarities)$ \Comment{Select most similar to centroid}
\State \Return $e_{best\_idx}$
\end{algorithmic}
\end{algorithm*}

\section{Additional Case Studies}
\label{appendix:case_studies}

Tables~\ref{tab:case_study_comve} to \ref{tab:case_study_esnli} present comprehensive case studies comparing NLF and IWF-Attn (our best-performing variant) refinement processes across three datasets. Table~\ref{tab:case_study_ecqa} provides the complete three-round refinement details for the ECQA example discussed in Section~\ref{sec:analysis}, while Tables~\ref{tab:case_study_comve} and \ref{tab:case_study_esnli} show representative examples from ComVE and e-SNLI datasets, respectively.

\begin{table*}[t]
\centering
\small
\begin{tabular}{p{7.5cm}|p{7.5cm}}
\toprule
\multicolumn{2}{l}{
  \begin{minipage}{\dimexpr 15cm + 2\tabcolsep + \arrayrulewidth\relax}
    \begin{tabular}{@{} p{2cm} p{13cm} @{}} 
      \textbf{Original} 
      & \textbf{Sentence 0:}  Leafs help plants absorb nutrition. \\
      & \textbf{Sentence 1:} The leafs are useless. \\
      & \textbf{Answer options:} (A) sentence 0 (B) sentence 1\\
      & \textbf{Model prediction:} (B) sentence 1
    \end{tabular}
  \end{minipage}
} \\
\midrule
\multicolumn{2}{l}{
  \begin{minipage}{\dimexpr 15cm + 2\tabcolsep + \arrayrulewidth\relax}
    \begin{tabular}{@{} p{2cm} p{13cm} @{}}
      \textbf{Intervened} 
      & \textbf{Sentence 0:}  Leafs help plants absorb nutrition. \\
      & \textbf{Sentence 1:} The \textcolor{blue}{\textbf{fallen}} leafs are useless. \\
      & \textbf{Answer options:} (A) sentence 0 (B) sentence 1\\
      & \textbf{Model prediction:} \textcolor{blue}{\textbf{(B) sentence 0}}
    \end{tabular}
  \end{minipage}
} \\
\midrule
\multicolumn{2}{l}{
  \begin{minipage}{\dimexpr 15cm + 2\tabcolsep + \arrayrulewidth\relax}
    \textbf{Initial Explanation:} Sentence 0 violates common sense because leaves do not help plants absorb nutrition; rather, it is the roots that absorb nutrients from the soil. \textcolor{red}{\textbf{Unfaithful}}
  \end{minipage}
} \\
\midrule
\multicolumn{1}{c}{\textbf{Natural Language Feedback (NLF)}} & \multicolumn{1}{c}{\textbf{Important-Word Feedback (IWF-Attn)}} \\
\midrule
\textbf{Feedback 1:} The explanation accurately reflects the reasoning behind selecting Sentence 0 as the one that violates common sense. It correctly identifies that leaves do not help plants absorb nutrition, but rather it is the roots. However, to improve the explanation, it could briefly mention that while leaves do not absorb nutrients, they play a crucial role in photosynthesis, which indirectly supports the plant's overall nutrition. Overall, the explanation is clear and concise. 
& \textbf{Feedback 1:} The 5 most important words that contributed to your prediction are: leafs, the, useless, fallen, are. \\
\midrule
\textbf{Refinement 1:} Sentence 0 violates common sense because leaves do not help plants absorb nutrition; instead, it is the roots that take up nutrients from the soil. Leaves, however, play a vital role in photosynthesis, which is essential for the plant's overall nutrition and growth.\textcolor{red}{\textbf{Unfaithful}}
& \textbf{Refinement 1:} Leaves do not assist in the absorption of nutrition; instead, it is the roots that play this crucial role. \textcolor{blue}{\textbf{Fallen}} leaves, however, can become useless as they decompose and return nutrients to the soil. \textcolor{green}{\textbf{Faithful}} \\
\midrule
\textbf{Feedback 2:} The explanation accurately reflects the reasoning behind selecting Sentence 0 as the one that violates common sense. It correctly identifies that leaves do not help plants absorb nutrition, but rather it is the roots. However, to improve the explanation, it could briefly mention that while leaves are not directly involved in nutrient absorption, they are crucial for photosynthesis, which indirectly supports the plant's nutrition. Overall, the explanation is clear and concise. 
& \textbf{Feedback 2:} The 5 most important words that contributed to your prediction are: leafs, the, useless, fallen, are. \\
\midrule
\textbf{Refinement 2:} Sentence 0 violates common sense because leaves do not help plants absorb nutrition; instead, it is the roots that take up nutrients from the soil. While leaves are not directly involved in nutrient absorption, they are essential for photosynthesis, which indirectly supports the plant's nutrition and overall growth. \textcolor{red}{\textbf{Unfaithful}}
& \textbf{Refinement 2:} Leaves do not assist in the absorption of nutrition; instead, it is the roots that play this crucial role. \textcolor{blue}{\textbf{Fallen}} leaves, however, can become useless as they decompose and return nutrients to the soil. \textcolor{green}{\textbf{Faithful}} \\
\midrule
\textbf{Feedback 3:} The explanation accurately reflects the reasoning behind selecting Sentence 0 as the one that violates common sense. It correctly identifies that leaves do not help plants absorb nutrition, and instead highlights the role of roots in nutrient absorption. However, to improve the explanation, it could briefly mention that while leaves are not directly involved in nutrient absorption, they play a crucial role in the overall health of the plant by facilitating photosynthesis, which indirectly supports the plant's nutrition. This addition would provide a more comprehensive understanding of the plant's nutritional processes.
& \textbf{Feedback 3:} The 5 most important words that contributed to your prediction are: leafs, the, useless, fallen, are. \\
\midrule
\textbf{Refinement 3:} Sentence 0 violates common sense because leaves do not help plants absorb nutrition; rather, it is the roots that take up nutrients from the soil. Leaves, however, are vital for photosynthesis, which indirectly supports the plant's nutrition and overall growth. \textcolor{red}{\textbf{Unfaithful}} 
& \textbf{Refinement 3:} Leaves do not assist in the absorption of nutrition; instead, it is the roots that play this crucial role. \textcolor{blue}{\textbf{Fallen}} leaves, however, can become useless as they decompose and return nutrients to the soil. \textcolor{green}{\textbf{Faithful}} \\
\bottomrule
\end{tabular}
\caption{Complete case study on the ComVE dataset. The intervened word, highlighted in \textcolor{blue}{\textbf{blue}}, successfully changes the model prediction. \textcolor{green}{\textbf{Faithful}} indicates the explanation/refinement is faithful as judged by the counterfactual test, while \textcolor{red}{\textbf{Unfaithful}} indicates the opposite.}
\label{tab:case_study_comve}
\end{table*}
\begin{table*}[t]
\centering
\small
\begin{tabular}{p{7.5cm}|p{7.5cm}}
\toprule
\multicolumn{2}{l}{
  \begin{minipage}{\dimexpr 15cm + 2\tabcolsep + \arrayrulewidth\relax}
    \begin{tabular}{@{} p{2cm} p{13cm} @{}} 
      \textbf{Original} 
      & \textbf{Question:} There was only one room in the place where Bill slept. It had a bed, a fridge, a stove, a couch, and a television. Where might he be? \\
      & \textbf{Answer options:} (A) motel (B) school (C) hotel (D) apartment (E) friend's house  \\
      & \textbf{Model prediction:} (D) apartment
    \end{tabular}
  \end{minipage}
} \\
\midrule
\multicolumn{2}{l}{
  \begin{minipage}{\dimexpr 15cm + 2\tabcolsep + \arrayrulewidth\relax}
    \begin{tabular}{@{} p{2cm} p{13cm} @{}}
      \textbf{Intervened} 
      & \textbf{Question:} There was only one \textcolor{blue}{\textbf{cozy}} room in the place where Bill slept. It had a bed, a fridge, a stove, a couch, and a television. Where might he be? \\
      & \textbf{Answer options:} (A) motel (B) school (C) hotel (D) apartment (E) friend's house  \\
      & \textbf{Model prediction:} \textcolor{blue}{\textbf{(A) motel}}
    \end{tabular}
  \end{minipage}
} \\
\midrule
\multicolumn{2}{l}{
  \begin{minipage}{\dimexpr 15cm + 2\tabcolsep + \arrayrulewidth\relax}
    \textbf{Initial Explanation:} Bill's room contains amenities typical of a motel, such as a bed, fridge, stove, couch, and television, which are not usually found together in a school setting. \textcolor{red}{\textbf{Unfaithful}}
  \end{minipage}
} \\
\midrule
\multicolumn{1}{c}{\textbf{Natural Language Feedback (NLF)}} & \multicolumn{1}{c}{\textbf{Important-Word Feedback (IWF-Attn)}} \\
\midrule
\textbf{Feedback 1:} The explanation accurately reflects the reasoning behind selecting option (A) as it highlights the motel-like amenities present in Bill's room. However, it could be improved by explicitly mentioning that the combination of a bed, fridge, stove, couch, and television is more characteristic of a motel or hotel room than other options like a school or apartment. This would make the reasoning clearer and more convincing. No other significant improvements are needed. 
& \textbf{Feedback 1:} The 5 most important words that contributed to your prediction are: one, a, cozy, be, there. \\
\midrule
\textbf{Refinement 1:} Bill's room contains a bed, fridge, stove, couch, and television, which are typical amenities found in a motel or hotel room, not in a school or apartment setting. This combination of features makes option (A) Motel the most appropriate choice. \textcolor{red}{\textbf{Unfaithful}}
& \textbf{Refinement 1:} Bill's room, described as \textcolor{blue}{\textbf{cozy}} with amenities like a bed, fridge, stove, couch, and television, suggests a motel setting, as these features are commonly found together in such accommodations. \textcolor{green}{\textbf{Faithful}} \\
\midrule
\textbf{Feedback 2:} The explanation accurately reflects the reasoning behind selecting option (A) Motel. It correctly identifies the presence of typical motel amenities and contrasts them with settings like schools and apartments. However, to enhance clarity, it could be beneficial to explicitly mention that motels are often the type of accommodation where travelers might find a combination of a bed, fridge, stove, couch, and television in a single room. This would further emphasize why option (A) is the most suitable choice. Overall, the explanation is mostly clear but could be slightly refined for better understanding. 
& \textbf{Feedback 2:} The 5 most important words that contributed to your prediction are: one, a, cozy, be, there. \\
\midrule
\textbf{Refinement 2:} Bill's room contains a bed, fridge, stove, couch, and television, which are typical amenities found in a motel or hotel room, not in a school or apartment setting. The presence of a bed, fridge, and stove suggests it's a place for temporary accommodation, and the inclusion of a couch and television indicates it's a comfortable space for relaxation. This combination of features makes option (A) Motel the most appropriate choice, as motels often provide such a suite of amenities in a single room for travelers. \textcolor{red}{\textbf{Unfaithful}}
& \textbf{Refinement 2:} Bill's \textcolor{blue}{\textbf{cozy}} room, equipped with a bed, fridge, stove, couch, and television, indicates a motel setting, where such amenities are typically found together. \textcolor{green}{\textbf{Faithful}} \\
\midrule
\textbf{Feedback 3:} The explanation accurately reflects the reasoning behind selecting option (A) Motel. It correctly identifies the amenities as typical of a motel room and contrasts them with other options. However, it could be improved by explicitly mentioning that the presence of a bed, fridge, and stove suggests a living space, which is more characteristic of a motel room than a school or apartment. No improvement is needed overall.
& \textbf{Feedback 3:} The 5 most important words that contributed to your prediction are: one, a, cozy, be, there. \\
\midrule
\textbf{Refinement 3:} Bill's room, equipped with a bed, fridge, stove, couch, and television, is indicative of a motel or hotel room, where such amenities are common for temporary accommodation and comfort. The presence of a bed, fridge, and stove suggests a living space, which aligns more with a motel setting than a school or apartment. This combination of features makes option (A) Motel the most appropriate choice. \textcolor{red}{\textbf{Unfaithful}} 
& \textbf{Refinement 3:} Bill's \textcolor{blue}{\textbf{cozy}} room, complete with a bed, fridge, stove, couch, and television, suggests a motel setting, where such a combination of amenities is commonly found. \textcolor{green}{\textbf{Faithful}} \\
\bottomrule
\end{tabular}
\caption{Complete case study on the ECQA dataset. The intervened word, highlighted in \textcolor{blue}{\textbf{blue}}, successfully changes the model prediction. \textcolor{green}{\textbf{Faithful}} indicates the explanation/refinement is faithful as judged by the counterfactual test, while \textcolor{red}{\textbf{Unfaithful}} indicates the opposite.}
\label{tab:case_study_ecqa}
\end{table*}
\begin{table*}[t]
\centering
\small
\begin{tabular}{p{7.5cm}|p{7.5cm}}
\toprule
\multicolumn{2}{l}{
  \begin{minipage}{\dimexpr 15cm + 2\tabcolsep + \arrayrulewidth\relax}
    \begin{tabular}{@{} p{2cm} p{13cm} @{}} 
      \textbf{Original} 
      & \textbf{Premise:} A guy riding a motorcycle near junk cars. \\
      & \textbf{Hypothesis:} A man is riding a motorcycle. \\      
      & \textbf{Answer options:} (A) contradiction (B) neutral (C) entailment  \\
      & \textbf{Model prediction:} (C) entailment
    \end{tabular}
  \end{minipage}
} \\
\midrule
\multicolumn{2}{l}{
  \begin{minipage}{\dimexpr 15cm + 2\tabcolsep + \arrayrulewidth\relax}
    \begin{tabular}{@{} p{2cm} p{13cm} @{}}
      \textbf{Intervened} 
      & \textbf{Premise:} A guy riding a motorcycle near junk cars. \\
      & \textbf{Hypothesis:} A man is riding a \textcolor{blue}{\textbf{powerful}} motorcycle. \\      
      & \textbf{Answer options:} (A) contradiction (B) neutral (C) entailment  \\
      & \textbf{Model prediction:} \textcolor{blue}{\textbf{(B) neutral}}
    \end{tabular}
  \end{minipage}
} \\
\midrule
\multicolumn{2}{l}{
  \begin{minipage}{\dimexpr 15cm + 2\tabcolsep + \arrayrulewidth\relax}
    \textbf{Initial Explanation:} The premise does not specify the power of the motorcycle, so the hypothesis introduces new information that cannot be confirmed from the premise. \textcolor{red}{\textbf{Unfaithful}}
  \end{minipage}
} \\
\midrule
\multicolumn{1}{c}{\textbf{Natural Language Feedback (NLF)}} & \multicolumn{1}{c}{\textbf{Important-Word Feedback (IWF-Attn)}} \\
\midrule
\textbf{Feedback 1:} The explanation accurately reflects the reasoning process and correctly identifies that the premise does not provide information about the motorcycle's power. However, it could be slightly more explicit by mentioning that the premise also does not confirm the presence of a man specifically. To improve clarity, you might add: "The premise does not specify the power of the motorcycle or confirm the presence of a man, so the hypothesis introduces new information that cannot be confirmed." 
& \textbf{Feedback 1:} The 5 most important words that contributed to your prediction are: a, near, powerful, is, guy. \\
\midrule
\textbf{Refinement 1:} The premise does not specify the power of the motorcycle or confirm the presence of a man, so the hypothesis introduces new information that cannot be confirmed. \textcolor{red}{\textbf{Unfaithful}}
& \textbf{Refinement 1:} The premise does not specify the power of the motorcycle, so the hypothesis introduces new information about the motorcycle being \textcolor{blue}{\textbf{powerful}}, which cannot be confirmed from the given premise. \textcolor{green}{\textbf{Faithful}} \\
\midrule
\textbf{Feedback 2:} The explanation accurately reflects the reasoning process. It correctly identifies that the premise does not provide information about the power of the motorcycle or confirm the presence of a man. However, it could be slightly more detailed by explicitly stating that the hypothesis introduces assumptions not supported by the premise. This minor clarification would make the explanation clearer. No significant improvements are needed, but adding this detail would enhance its precision.
& \textbf{Feedback 2:} The 5 most important words that contributed to your prediction are: a, near, powerful, is, guy. \\
\midrule
\textbf{Refinement 2:} The premise does not specify the power of the motorcycle or confirm the presence of a man, so the hypothesis introduces assumptions not supported by the premise. \textcolor{red}{\textbf{Unfaithful}}
& \textbf{Refinement 2:} The premise does not specify the power of the motorcycle, so the hypothesis introduces new information about the motorcycle being \textcolor{blue}{\textbf{powerful}}, which cannot be confirmed from the given premise. \textcolor{green}{\textbf{Faithful}} \\
\midrule
\textbf{Feedback 3:} The explanation accurately reflects the reasoning process. It correctly identifies that the premise does not provide information about the power of the motorcycle or confirm the presence of a man. However, it could be slightly more explicit about why the hypothesis is considered neutral rather than just providing a general statement. A minor revision could clarify this point. Revised Explanation: The premise does not specify the power of the motorcycle or confirm the presence of a man. Therefore, the hypothesis introduces assumptions not supported by the premise, making the relationship neutral. Overall, the explanation is good but can be made clearer with a slight revision.
& \textbf{Feedback 3:} The 5 most important words that contributed to your prediction are: a, near, powerful, is, guy. \\
\midrule
\textbf{Refinement 3:} The premise does not specify the power of the motorcycle or confirm the presence of a man. Therefore, the hypothesis introduces assumptions not supported by the premise, making the relationship neutral. \textcolor{red}{\textbf{Unfaithful}} 
& \textbf{Refinement 3:} The premise does not specify the power of the motorcycle, so the hypothesis introduces new information about the motorcycle being \textcolor{blue}{\textbf{powerful}}, which cannot be confirmed from the given premise. \textcolor{green}{\textbf{Faithful}} \\
\bottomrule
\end{tabular}
\caption{Complete case study on the e-SNLI dataset. The intervened word, highlighted in \textcolor{blue}{\textbf{blue}}, successfully changes the model prediction. \textcolor{green}{\textbf{Faithful}} indicates the explanation/refinement is faithful as judged by the counterfactual test, while \textcolor{red}{\textbf{Unfaithful}} indicates the opposite.}
\label{tab:case_study_esnli}
\end{table*}


\section{Prompts}
\label{appendix:prompt}

We present the full prompt templates used in our experiments. Each prompt is composed of two parts: a task-specific prompt part and a common instruction part, which are concatenated to form the final prompt at each stage of the SR-NLE framework. These stages include answer generation, explanation generation, feedback generation, and refinement generation. For feedback and refinement, we provide two variants based on natural language feedback and important word feedback. Tables~\ref{tab:prompt-answer} to \ref{tab:prompt-refinement-iw} list the complete prompts for each stage, including the task-specific prompt part for all three datasets and the shared common instruction part.

\begin{table*}[ht]
\centering
\begin{tabular}{p{0.10\linewidth} | p{0.40\linewidth} | p{0.40\linewidth}}
\toprule
\textbf{Dataset} & \textbf{Task Specific Prompt Part} & \textbf{Common Instruction Prompt Part} \\
\midrule
\bf ComVE &
You are given two sentences. Identify which one violates commonsense.\newline
\newline
Sentence 0: \{sentence0\}\newline
Sentence 1: \{sentence1\}\newline
Answer Options:\newline
(A) Sentence 0\newline
(B) Sentence 1
&
\multirow{3}{=}{Please select the most appropriate answer without any explanation.\newline
\newline
You must give your answer only in the following format:\newline
Answer: (X)
}
\\
\cmidrule{1-2}
\bf ECQA &
You are given a multiple-choice commonsense question. Identify the most appropriate answer.\newline
\newline
Question: \{question\}\newline
Answer Options:\newline
(A) \{Option 1\}\newline
(B) \{Option 2\}\newline
(C) \{Option 3\}\newline
(D) \{Option 4\}\newline
(E) \{Option 5\}
&
\\
\cmidrule{1-2}
\bf e-SNLI &
You are given a premise and a hypothesis. Identify the logical relationship between them.\newline
Premise: \{premise\}\newline
Hypothesis: \{hypothesis\}\newline
Answer Options:\newline
(A) Contradiction\newline
(B) Neutral\newline
(C) Entailment
&
\\
\bottomrule
\end{tabular}
\caption{Answer generation prompts.}
\label{tab:prompt-answer}
\end{table*}

\begin{table*}[ht]
\centering
\begin{tabular}{p{0.10\linewidth} | p{0.40\linewidth} | p{0.40\linewidth}}
\toprule
\textbf{Dataset} & \textbf{Task Specific Prompt Part} & \textbf{Common Instruction Prompt Part} \\
\midrule
\bf ComVE &
You are given two sentences, and you have selected the one that violates commonsense.\newline
\newline
Sentence 0: \{sentence0\}\newline
Sentence 1: \{sentence1\}\newline
Answer Options:\newline
(A) Sentence 0\newline
(B) Sentence 1
&
\multirow{3}{=}{Your selected answer is: ([LABEL]).\newline
\newline
Now, please provide an explanation for your choice.\newline
\newline
Your explanation should:\newline
- Be clear, complete, and concise.\newline
- Ideally within two short sentences.\newline
\newline
You must give your explanation only in the following format:\newline
Explanation: [your explanation here.]\newline
}
\\
\cmidrule{1-2}
\bf ECQA &
You are given a multiple-choice commonsense question, and you have selected the most appropriate answer.\newline
\newline
Question: \{question\}\newline
Answer Options:\newline
(A) \{Option 1\}\newline
(B) \{Option 2\}\newline
(C) \{Option 3\}\newline
(D) \{Option 4\}\newline
(E) \{Option 5\}
&
\\
\cmidrule{1-2}
\bf e-SNLI &
You are given a premise and a hypothesis, and you have selected the logical relationship between them.\newline
\newline
Premise: \{premise\}\newline
Hypothesis: \{hypothesis\}\newline
Answer Options:\newline
(A) Contradiction\newline
(B) Neutral\newline
(C) Entailment
&
\\
\bottomrule
\end{tabular}
\caption{Explanation generation prompts.}
\label{tab:prompt-explanation}
\end{table*}

\begin{table*}[ht]
\centering
\begin{tabular}{p{0.10\linewidth} | p{0.40\linewidth} | p{0.40\linewidth}}
\toprule
\textbf{Dataset} & \textbf{Task Specific Prompt Part} & \textbf{Common Instruction Prompt Part} \\
\midrule
\bf ComVE &
You are given two sentences, and you have selected the one that violates commonsense. You then provided an explanation for your choice.\newline
\newline
Sentence 0: \{sentence0\}\newline
Sentence 1: \{sentence1\}\newline
Answer Options:\newline
(A) Sentence 0\newline
(B) Sentence 1
&
\multirow{3}{=}{Your selected answer is: ([LABEL])\newline
Your explanation is:\newline
[EXPLANATION]\newline
\newline
Now, please provide feedback on this explanation.\newline
\newline
Your feedback should:\newline
- Identify whether the explanation accurately reflects your actual reasoning.\newline
- Point out if any key factors or important details are missing, unclear, or incorrect.\newline
- Briefly describe what should be added or revised to improve the explanation.\newline
- Clearly state that no improvement is needed when the explanation is good enough.\newline
- Be concise, avoid unnecessary repetition or irrelevant details.\newline
\newline
You must give your feedback only in the following format:\newline
Feedback: [your feedback here.]\newline
}
\\
\cmidrule{1-2}
\bf ECQA &
You are given a multiple-choice commonsense question, and you have selected the most appropriate answer. You then provided an explanation for your choice.\newline
\newline
Question: \{question\}\newline
Answer Options:\newline
(A) \{Option 1\}\newline
(B) \{Option 2\}\newline
(C) \{Option 3\}\newline
(D) \{Option 4\}\newline
(E) \{Option 5\}
&
\\
\cmidrule{1-2}
\bf e-SNLI &
You are given a premise and a hypothesis, and you have selected the logical relationship between them. You then provided an explanation for your choice.\newline
\newline
Premise: \{premise\}\newline
Hypothesis: \{hypothesis\}\newline
Answer Options:\newline
(A) Contradiction\newline
(B) Neutral\newline
(C) Entailment
&
\\
\bottomrule
\end{tabular}
\caption{Natural language feedback generation prompts.}
\label{tab:prompt-feedback-nl}
\end{table*}

\begin{table*}[ht]
\centering
\begin{tabular}{p{0.10\linewidth} | p{0.40\linewidth} | p{0.40\linewidth}}
\toprule
\textbf{Dataset} & \textbf{Task Specific Prompt Part} & \textbf{Common Instruction Prompt Part} \\
\midrule
\bf ComVE &
You are given two sentences, and you have selected the one that violates commonsense.\newline
\newline
Sentence 0: \{sentence0\}\newline
Sentence 1: \{sentence1\}\newline
Answer Options:\newline
(A) Sentence 0\newline
(B) Sentence 1
&
\multirow{3}{=}{Your selected answer is: ([LABEL]).\newline
\newline
Now, please evaluate all the words in the input and rank them by how important they were in helping you make your choice.\newline
\newline
Your output must meet the following requirements:\newline
- Only include individual words in the input.\newline
- Evaluate each word based on its total contribution across all occurrences in the input, but include each word only once in the output.\newline
- Assign each word a score from 1 to 100 (positive integers only), based on its relative importance.\newline
- Rank the words in descending order of importance (most important first).\newline
- Do not include any explanations, comments, or parenthetical notes.\newline
\newline
You must give your output only in the following format:\newline
- Begin directly with the ranked list.\newline
- Each line must be in the format:\newline
  `<rank>. <word>, <importance\_score>`\newline
}
\\
\cmidrule{1-2}
\bf ECQA &
You are given a multiple-choice commonsense question, and you have selected the most appropriate answer.\newline
\newline
Question: \{question\}\newline
Answer Options:\newline
(A) \{Option 1\}\newline
(B) \{Option 2\}\newline
(C) \{Option 3\}\newline
(D) \{Option 4\}\newline
(E) \{Option 5\}
&
\\
\cmidrule{1-2}
\bf e-SNLI &
You are given a premise and a hypothesis, and you have selected the logical relationship between them.newline\newline
\newline
Premise: \{premise\}\newline
Hypothesis: \{hypothesis\}\newline
Answer Options:\newline
(A) Contradiction\newline
(B) Neutral\newline
(C) Entailment
&
\\
\bottomrule
\end{tabular}
\caption{Important words feedback generation prompts.}
\label{tab:prompt-feedback-iw}
\end{table*}

\begin{table*}[ht]
\centering
\begin{tabular}{p{0.10\linewidth} | p{0.40\linewidth} | p{0.40\linewidth}}
\toprule
\textbf{Dataset} & \textbf{Task Specific Prompt Part} & \textbf{Common Instruction Prompt Part} \\
\midrule
\bf ComVE &
You are given two sentences, and you have selected the one that violates commonsense. You then provided an explanation for your choice, and received feedback on the explanation.\newline
\newline
Sentence 0: \{sentence0\}\newline
Sentence 1: \{sentence1\}\newline
Answer Options:\newline
(A) Sentence 0\newline
(B) Sentence 1
&
\multirow{3}{=}{Your selected answer is: ([LABEL])\newline
Your explanation is:\newline
[EXPLANATION]\newline
The feedback you received is:\newline
[FEEDBACK]\newline
\newline
If the feedback indicates that no improvement is needed, you should repeat the original explanation as the refined explanation.
Otherwise, please refine your explanation based on the feedback.\newline
\newline
Your refined explanation should:\newline
- Be clear, complete, and concise.\newline
- Ideally remain similar in length to the original explanation.\newline
- Retain any correct parts of your original explanation.\newline
- Address the issues identified in the feedback, if any.\newline
\newline
You must give your refined explanation only in the following format:\newline
Refined Explanation: [your refined explanation here.]\newline
}
\\
\cmidrule{1-2}
\bf ECQA &
You are given a multiple-choice commonsense question, and you have selected the most appropriate answer. You then provided an explanation for your choice, and received feedback on the explanation.\newline
\newline
Question: \{question\}\newline
Answer Options:\newline
(A) \{Option 1\}\newline
(B) \{Option 2\}\newline
(C) \{Option 3\}\newline
(D) \{Option 4\}\newline
(E) \{Option 5\}
&
\\
\cmidrule{1-2}
\bf e-SNLI &
You are given a premise and a hypothesis, and you have selected the logical relationship between them. You then provided an explanation for your choice, and received feedback on the explanation.\newline
\newline
Premise: \{premise\}\newline
Hypothesis: \{hypothesis\}\newline
Answer Options:\newline
(A) Contradiction\newline
(B) Neutral\newline
(C) Entailment
&
\\
\bottomrule
\end{tabular}
\caption{Refinement generation prompts based on natural language feedback.}
\label{tab:prompt-refinement-nl}
\end{table*}

\begin{table*}[ht]
\centering
\begin{tabular}{p{0.10\linewidth} | p{0.40\linewidth} | p{0.40\linewidth}}
\toprule
\textbf{Dataset} & \textbf{Task Specific Prompt Part} & \textbf{Common Instruction Prompt Part} \\
\midrule
\bf ComVE &
You are given two sentences, and you have selected the one that violates commonsense. You then provided an explanation for your choice, and received a list of important words that contributed significantly to your reasoning.\newline
\newline
Sentence 0: \{sentence0\}\newline
Sentence 1: \{sentence1\}\newline
Answer Options:\newline
(A) Sentence 0\newline
(B) Sentence 1
&
\multirow{3}{=}{Your selected answer is: ([LABEL])\newline
Your explanation is:\newline
[EXPLANATION]\newline
The important words you received are:\newline
[FEEDBACK]\newline
\newline
If the explanation already includes the important words in a natural and meaningful way, you should repeat the original explanation as the refined explanation. Otherwise, please refine your explanation based on the important words.\newline
\newline
Your refined explanation should:\newline
- Be clear, complete, and concise.\newline
- Ideally remain similar in length to the original explanation.\newline
- Retain any correct parts of your original explanation.\newline
- Integrate the important words naturally and fluently—do not list or quote them directly.\newline
\newline
Provide your refined explanation only in the following format:\newline
Refined Explanation: [your refined explanation here.]\newline
}
\\
\cmidrule{1-2}
\bf ECQA &
You are given a multiple-choice commonsense question, and you have selected the most appropriate answer. You then provided an explanation for your choice, and received a list of important words that contributed significantly to your reasoning.\newline
\newline
Question: \{question\}\newline
Answer Options:\newline
(A) \{Option 1\}\newline
(B) \{Option 2\}\newline
(C) \{Option 3\}\newline
(D) \{Option 4\}\newline
(E) \{Option 5\}
&
\\
\cmidrule{1-2}
\bf e-SNLI &
You are given a premise and a hypothesis, and you have selected the logical relationship between them. You then provided an explanation for your choice, and received a list of important words that contributed significantly to your reasoning.\newline
\newline
Premise: \{premise\}\newline
Hypothesis: \{hypothesis\}\newline
Answer Options:\newline
(A) Contradiction\newline
(B) Neutral\newline
(C) Entailment
&
\\
\bottomrule
\end{tabular}
\caption{Refinement generation prompts based on important words feedback.}
\label{tab:prompt-refinement-iw}
\end{table*}


\end{document}